\pdfoutput=1

\documentclass[11pt]{article}

\usepackage{EMNLP2023}

\usepackage{times}
\usepackage{latexsym}
\usepackage{amssymb}

\usepackage{amsmath}
\usepackage{graphicx}
\usepackage{makecell}
\usepackage{booktabs}
\usepackage{enumitem}
\usepackage{adjustbox}

\usepackage{relsize}

\usepackage{amsmath}
\DeclareMathOperator*{\argmax}{argmax} 

\usepackage{multirow}
\usepackage{siunitx}

\usepackage[T1]{fontenc}

\usepackage[utf8]{inputenc}

\usepackage{microtype}

\usepackage{inconsolata}

\usepackage{amssymb}
\usepackage{pifont}

\newcommand{\halluNE}{$\textsc{NE}_{\textsc{Er}}$}
\newcommand{\entailedRatio}{$\text{Entail}_{\textsc{R}}$}

\makeatletter
\renewcommand{\@fnsymbol}[1]{\ensuremath{\ifcase#1\or \dagger\or \ddagger\or \S\or \P\or \parallel\or **\or \dagger\dagger\or \ddagger\ddagger\else\@ctrerr\fi}}
\makeatother

\title{Beyond Factuality: A Comprehensive Evaluation of Large Language Models as Knowledge Generators}

\author{
Liang Chen\textsuperscript{1}, Yang Deng\textsuperscript{2}, Yatao Bian\textsuperscript{3}\thanks{ \hspace{1mm} Corresponding author.}, 
Zeyu Qin\textsuperscript{4}, 
\\
\textbf{
Bingzhe Wu\textsuperscript{3}, 
Tat-Seng Chua\textsuperscript{2}, Kam-Fai Wong\textsuperscript{1}\footnotemark[1]}  \\
\textsuperscript{1}The Chinese University of Hong Kong, \textsuperscript{2}National University of Singapore,
  \textsuperscript{3}Tencent AI Lab \\ 
  \textsuperscript{4}The Hong Kong University of Science and Technology  \\
  \texttt{lchen@se.cuhk.edu.hk} 
}

\begin{document}
\maketitle

\begin{abstract}
Large language models (LLMs) outperform information retrieval techniques for downstream knowledge-intensive tasks when being prompted to generate world knowledge.
Yet, community concerns abound regarding the factuality and potential implications of using this uncensored knowledge.
In light of this, we introduce \texttt{CONNER}, a COmpreheNsive kNowledge Evaluation fRamework, designed to systematically and automatically evaluate generated knowledge from six 
important perspectives -- \textit{Factuality, Relevance}, \textit{Coherence}, \textit{Informativeness}, \textit{Helpfulness} and \textit{Validity}.
We conduct an extensive empirical analysis of the generated knowledge from three different types of LLMs on two widely-studied knowledge-intensive tasks, \textit{i.e.}, open-domain question answering and knowledge-grounded dialogue. 
Surprisingly, our study reveals that the factuality of generated knowledge, even if lower, does not significantly hinder downstream tasks. Instead, the relevance and coherence of the outputs are more important than small factual mistakes.
Further, we show how to use \texttt{CONNER} to improve knowledge-intensive tasks by designing two strategies: Prompt Engineering and Knowledge Selection.
Our evaluation code and LLM-generated knowledge with human annotations will be released\footnote{\url{https://github.com/ChanLiang/CONNER}} to facilitate future research.

\end{abstract}

\begin{figure}
    \centering
    \includegraphics[width=\linewidth]{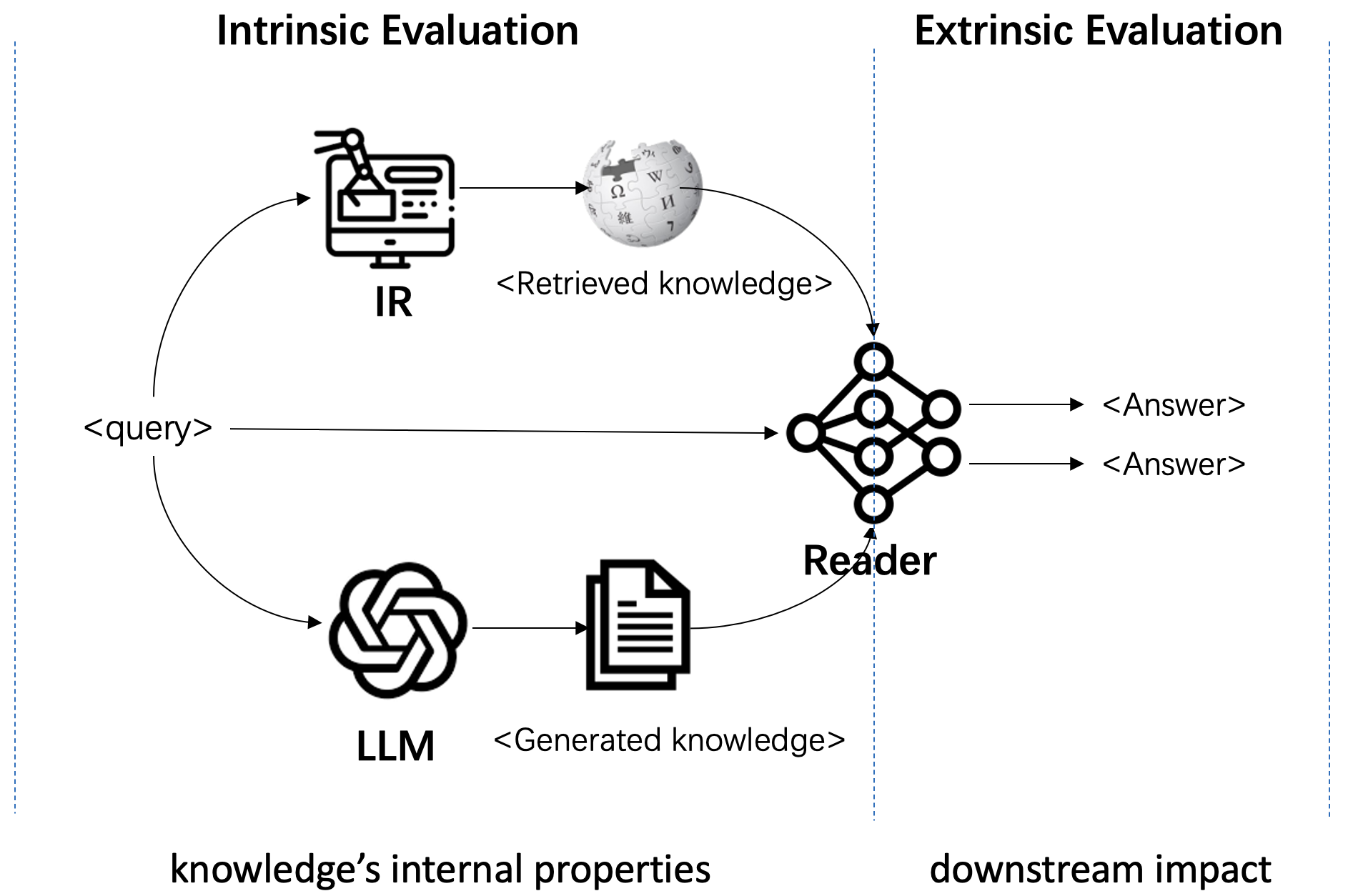}
    \caption{The \texttt{CONNER} Framework: Intrinsic evaluations probe the internal properties of acquired knowledge, while extrinsic evaluations assess its downstream impacts. This framework applies universally to two-stage processes in knowledge-intensive tasks.}
    \label{fig:framework}
  \vspace{-0.58cm}
\end{figure}

\section{Introduction}

         
       

        
        


\begin{table*}[t!]
\setlength{\abovecaptionskip}{5pt}   
\setlength{\belowcaptionskip}{0pt}
    \centering
    {\small
    \begin{tabular}{ll|p{10.8cm}}
        \toprule

        \multicolumn{2}{l|}{\makecell[c]{\textbf{Evaluation Taxonomy}}} & \makecell[c]{\textbf{Definition}} \\
        
        \midrule
        \multirow{4}{*}{\emph{Intrinsic}} & \texttt{Factuality} & whether the information in the knowledge can be verified by external evidence. \\
        
        &\texttt{Relevance} & whether the knowledge is relevant to the user query. \\
        
        &\texttt{Coherence} & whether the knowledge is coherent at the sentence and paragraph levels. \\

        &\texttt{Informativeness} & whether the knowledge is new or unexpected against the model's existing knowledge. \\

        \midrule
        \multirow{2}{*}{\emph{Extrinsic}} &
        \texttt{Helpfulness} & whether the knowledge can improve the downstream tasks. \\

        &\texttt{Validity} & whether the results of downstream tasks using the knowledge are factually accurate.  \\
        
        \bottomrule
    \end{tabular}
    }
    \caption{Taxonomy of evaluation metrics of acquired knowledge.}
    \label{tab:taxonomy}
    \vspace{-0.4cm}
\end{table*}

The exceptional success of large language models (LLMs) like ChatGPT and GPT4~\cite{ouyang2022training,gpt4} has fueled a growing interest in substituting traditional models with LLMs to attain superior performance across various NLP tasks~\cite{gpt_sum,query_expansion,gpt_ner}.
In open-domain question answering (QA) and knowledge-grounded dialogue, LLMs have demonstrated superior performance than information retrieval (IR) models \cite{karpukhin2020dense} when it comes to generating world knowledge \cite{yu2023generate,liu2022multistage} for the downstream tasks. 
However, the knowledge generated may contain inherent issues, such as false statements or off-topic information.
Therefore, the lack of extensive evaluation of this knowledge raises concerns about its use in downstream tasks.

To this end, four lines of research emerge. 
Firstly, human evaluations are conducted to assess the generated knowledge from diverse perspectives \cite{emnlp22-kgc,yu2023generate,liu2023evaluating}. 
However, their time-consuming nature and subjectivity often encounter issues of scalability and reproducibility.
Secondly, datasets have been constructed to evaluate open-domain generation with the aid of references \cite{Q2,revisit_nli,lee2023factuality,halueval}. 
These methods, while more objective, are limited by their dependence on human-labelled references, impacting their real-world applicability and generalizability to dynamically generated content.
Thirdly, self-evaluation methods \cite{LM_know,self_consistency} estimate a model's uncertainty in its generated content. Despite simplicity, they lack interpretability and are less effective for long-form answers.
Lastly, contemporary studies \cite{pan2023factchecking,min2023factscore} apply fact-checking principles to spot factual inaccuracies.
However, these evaluation methods mainly assess a single aspect of the intrinsic quality of generated knowledge, overlooking other facets and their extrinsic impact on downstream tasks, thereby limiting a comprehensive understanding of LLM-generated content.

In light of these limitations, we propose \texttt{CONNER}, a COmpreheNsive kNowledge Evaluation fRamework, as illustrated in Figure \ref{fig:framework}.
\texttt{CONNER} is designed to be a reference-free framework that can systematically and automatically evaluate the generated knowledge from six fine-grained perspectives, including diverse intrinsic evaluation of its internal properties, as well as uniform extrinsic evaluation of its impact on specific downstream tasks. 
The taxonomy of evaluation metrics is presented in Table~\ref{tab:taxonomy}. 
Based on \texttt{CONNER}, we conduct empirical evaluations on three different types of LLMs, including LLaMA \cite{wei2022finetuned} (a base LLM), FLAN-T5 \cite{wei2022finetuned} (an instruction-tuned LLM), ChatGPT \cite{ouyang2022training} (a commercial LLM trained with human feedbacks). We evaluate them on two widely-studied knowledge-intensive tasks: open-domain QA \cite{kwiatkowski-etal-2019-natural} and knowledge-grounded dialogue \cite{dinan2018wizard}.

Our detailed investigations yield several valuable insights about the LLM-generated knowledge: 1) LLM-generated knowledge surpasses retrieved knowledge in most evaluation perspectives, while it actually suffers from the factuality issue as expected. Notably, the factuality of downstream tasks is found to be less affected by this issue, when compared to the impact of lower relevancy and coherency observed in the retrieved knowledge ($\S$ \ref{sec:eval}).
2) Several critical factors are identified to influence the factuality of the generated knowledge, such as their frequency and length, while few-shot in-context learning and larger size of models do not necessarily guarantee higher quality and reliability ($\S$ \ref{sec:analysis}). 
3) In addition to assessing and analyzing the generated knowledge from different LLMs, the evaluation outcome of \texttt{CONNER} can be exploited to enhance knowledge generation and further improve the performance of downstream tasks ($\S$ \ref{improvement}). 

Our main contributions are as follows: 
\vspace{-2.4mm}
\begin{itemize}[leftmargin=*]
    \item We conduct the first empirical analysis focusing on both intrinsic quality and extrinsic reliability of the generated knowledge from LLMs. \vspace{-2.2mm}
    \item We propose \texttt{CONNER}, a COmpreheNsive kNowledge Evaluation fRamework that enables the automatic evaluation of LLMs as knowledge generators from diverse perspectives, eliminating the need for human-labelled references. \vspace{-2.2mm}
    \item The extensive evaluation and analysis yield profound insights and valuable practical experience for leveraging LLMs as knowledge generators. \vspace{-2.2mm}
    \item We collect a new set of multi-perspective human judgments of LLM-generated knowledge for two knowledge-intensive generation datasets. We demonstrate that \texttt{CONNER} aligns well with human judgments. The human annotations will be released to facilitate future research. 
\end{itemize}

\section{Related Work}

Knowledge-intensive tasks rely heavily on access to external knowledge sources, such as open-domain dialogue and QA \cite{dinan2018wizard, kwiatkowski-etal-2019-natural,petroni2021kilt}. The main-streamed methods \cite{karpukhin2020dense,nips20-rag,izacard2021leveraging,acl23-esc} typically employ IR techniques to first retrieve the relevant knowledge from Wikipedia and then produce the answer or response conditioned on the knowledge. 
Nowadays, with the powerful capabilities of LLMs~\cite{gpt4,kadavath2022language}, a new trending approach is to leverage LLMs to directly generate the relevant knowledge for a given query and then apply the model-generated knowledge to complete the downstream tasks \cite{liu2022multistage,emnlp22-kgc,yu2023generate}.  
Despite the better performance than retrieval-based methods, there is a lack of rigorous evaluation of the quality and reliability of the generated knowledge, which may contain misleading or even plausible false information, \textit{e.g.}, hallucination and factual inconsistency. 

These issues are prevalent 
across various NLP tasks \cite{hallucination_survey_2023}. However, most studies target specific downstream tasks, such as text summarization~\cite{acl20-summ-fact,QGQA,factcc,pagnoni-etal-2021-understanding}, dialogue generation~\cite{shuster-etal-2021-retrieval-augmentation, dziri2022faithdial, chen-etal-2023-towards-robust,acl23-tutorial}, and fact verification \cite{fever,wadden-etal-2020-fact,schuster-etal-2021-get,pan2023factchecking}. These tasks are designed to examine consistency either between the input and output or between the input and a human-labeled reference, \textit{e.g.,} the source document and its summary, the grounded knowledge and the generated response, or a human-written claim and pre-annotated references.

The success of LLMs and generative search engines have brought hallucinations in LLM outputs \cite{zhang2023sirens} into focus. Research typically falls into four categories. \cite{lee2023factuality,halueval} aim to assess the factuality of open-domain generation automatically using specially designed datasets, but their reliance on references may limit real-world applicability. Another stream of work \cite{emnlp22-kgc,yu2023generate,liu2023evaluating} uses human evaluation to measure output quality, which is difficult to scale. A third approach \cite{LM_know,self_consistency} detects hallucinations by examining the model's uncertainty or confidence, which can be inaccurate for long answers. Lastly, recent studies \cite{peng2023check, pan2023factchecking,min2023factscore} apply fact-checking principles to spot factual inaccuracies.

Different from previous studies, we propose a \textbf{comprehensive framework} for evaluating knowledge generated by LLMs. Our goal is to \textbf{automatically test} the intrinsic quality and extrinsic impact of generated information in knowledge-intensive tasks, \textit{without requiring knowledge labelling or human involvement}. Through extensive testing with this framework, we aim to deepen and broaden our understanding of LLM-generated knowledge and provide valuable insights for future research.

\section{The Evaluation Framework}
We introduce \texttt{CONNER}, a comprehensive and innovative framework, specifically designed for the rigorous evaluation of the quality and dependability of knowledge used in knowledge-intensive tasks. \texttt{CONNER} is rooted in in-depth error analysis, paving the way for the construction of an evaluation taxonomy, which integrates six unique perspectives into two coherent categories, as delineated in Table~\ref{tab:taxonomy}. Capitalizing on the advantages of unsupervised metrics, our framework eliminates the need for human-labeled reference knowledge and standardizes scores within an intuitive range of $[0, 1]$, simplifying comparison and interpretation.

The subsequent subsections provide a detailed examination of the framework's design, commencing with the formulation of knowledge-intensive tasks and the identification of associated error patterns. These insights direct the design of our metrics. Through comprehensive intrinsic and extrinsic evaluations, 
we aim to gain a holistic understanding of the LLMs-generated knowledge.

\subsection{Tasks Formulation}

Formally, we define the knowledge-intensive task as follows: given a user query $q$, the goal is to produce an answer with access to knowledge resources.
Specifically, the system first obtains the relevant knowledge $k$ that can help answer the query $q$ from knowledge resources $\mathcal{K}$,
then generate an answer $a$ using the acquired knowledge $k$. 
Specifically, the knowledge resource $\mathcal{K}$ can be either a knowledge base for knowledge retrieval or language models for knowledge generation. Detailed formulations of these two settings are presented in Appendix~\ref{app:problem}.

\subsection{From Error Patterns to Metrics Design}
To identify common errors by LLMs in knowledge-intensive tasks and create a more targeted evaluation framework, we used thematic analysis~\cite{theme_ana}. 
We began by extracting and consolidating patterns from subtle errors in knowledge and answers in responses from LLaMA to 160 samples from NQ \cite{kwiatkowski-etal-2019-natural} and WoW \cite{dinan2018wizard} datasets. 
To ensure the breadth of the error spectrum was adequately represented, we further substantiated these patterns using additional questions from NQ and WoW.
As a result, we discerned
four primary error categories in knowledge generation and two in answer generation. In response, we devised four intrinsic metrics for knowledge evaluation and two extrinsic metrics for answer evaluation, as outlined in Table~\ref{tab:taxonomy}. 

\subsection{Intrinsic Evaluation}
Intrinsic evaluation refers to the assessment of the acquired knowledge based on its internal properties and performance, without considering its impact on downstream tasks or applications. In specific, we implement four model-based metrics for evaluating the acquired knowledge in terms of \textit{factuality}, \textit{relevance}, \textit{informativeness}, and \textit{coherence}.

\paragraph{Factuality} The core of factuality assessment is validating the acquired knowledge by external evidence
\footnote{We empirically demonstrate ground-truth knowledge is dispensable for the factuality evaluation in Appendix~\ref{app:ground-truth}.}. 
Given an acquired knowledge $k=\{s_1,\dots,s_m\}$ composed of $m$ sentences, we can use a dense retrieval model \cite{colbertv2} or search engine API to recall the $l_i$ most relevant evidence $E_i=\{e_{i,1},\dots,e_{i,l_i}\}$ for each sentence $s_i$ from the expert knowledge base or the internet. After collecting all the evidence $E=\{E_{1},\dots,E_{m}\}$, the factuality score is computed as follows:
\begin{equation}\small
\begin{split}
 \mathbf{S_\texttt{fact}}(k, E)&=\min_{i=1..m}f(s_i, E_i) \\ 
 &=\min_{i=1..m}\max_{j=1..l_i}\mathrm{NLI}(s_i, e_{i,j}) \label{eq:factuality}
 \end{split}
\end{equation} 
where $f(\cdot)$ is a function to compute sentence-level factuality, $\mathrm{NLI}(\cdot)$ is a natural language inference model processing a premise-hypothesis pair to output a $R^3$ vector, indicating whether a hypothesis ($s_i$) is entailed by, neutral to or refuted by the given premise ($e_{i,j}$). Following these computations, sentence-level results are aggregated along the entailment dimension using one of three operations: $\mathrm{min}$, $\mathrm{mean}$, or $\mathrm{max}$ to match the desired error tolerance level. In this instance, we exemplify the process using $\mathrm{min}$.
Finally, we obtain a three-dimensional factuality score $\mathbf{S_\texttt{fact}}(k, E)$. From each dimension of this vector, we can derive three fine-grained scores.
We denote those scores as \texttt{factual-consistent}, \texttt{non-verified}, and \texttt{factual-inconsistent}, respectively.

This strategy seeks to address the shortcomings of traditional factuality metrics \cite{QGQA,Q2,revisit_nli,lee2023factuality} that mainly depend on consistency with human-annotated references. These metrics often fail in emerging knowledge generation scenarios (Table \ref{tab:bl_col}), as they struggle with model-generated content beyond reference knowledge scope and face difficulties when references are unavailable in real-world applications. Our method of evidence collection and results aggregation effectively tackles these issues.

\paragraph{Relevance} To assess the relevance between a given query $q$ and the acquired knowledge $k$, we compute the relevance score as follows:
\begin{equation}\small
 S_\texttt{rel}(k, q)=\mathrm{Matching}(k, q) 
\end{equation} 
The $\mathrm{Matching}(\cdot)$ function denotes a fine-grained matching model specifically designed for assessing the relevance between the query and knowledge. In our study, we employ the $\mathrm{BERT}$ ranking model \cite{bert_ranker} for this purpose. 

This methodology addresses the limitations that arise when traditional relevance metrics are applied within knowledge generation scenarios. Traditional relevance metrics \cite{karpukhin2020dense,shuster-etal-2021-retrieval-augmentation,komeili2021internetaugmented}, which typically rely on word overlap or similarity with human-written references, face two significant challenges. First, these traditional metrics do not correspond well with scenarios where LLMs serve as generative search engines, as evidenced by the unsatisfactory results in Table \ref{tab:bl_col}. Second, the reliance on reference knowledge constitutes a substantial challenge, especially when such references are scarce or absent in real-world applications. Contrarily, our BERT ranking model, trained on manually annotated Bing search data, excels at comparing the relevance of different knowledge to a given query.

\paragraph{Coherence} 
As the acquired knowledge is typically long-form texts composed of multiple sentences, we propose to measure sentence-level cohesion and paragraph-level coherence: the former measures the cohesion of individual sentences, and the latter measures the coherence between sentences.
The sentence-level cohesion score $ S_\texttt{coh\_sent}(k) $ is computed as follows:
\begin{equation}\small
 S_\texttt{coh\_sent}(k)= \frac{1}{m} \sum\nolimits_{i=1}^{m} 1/\mathrm{PPL}(s_i)
\end{equation}
where $\mathrm{PPL}(\cdot)$ is computed by a GPT-based model \cite{gpt2,gpt-neo}, measuring the perplexity for each sentence.

On the other hand, the paragraph-level coherence score is determined by the normalized score of a discourse coherence model~\cite{para_coh}, denoted as $S_\texttt{coh\_para}(k)$:
\begin{equation}\small
  S_\texttt{coh\_para}(k)=\mathrm{\mathrm{Scorer}_{para}}(s_1,...,s_{m})
\end{equation}
By considering both sentence-level cohesion and paragraph-level coherence, we gain insights into the overall coherence of the acquired knowledge.

\paragraph{Informativeness} 

To assess the informativeness of the procured knowledge—defined as the degree to which the knowledge is novel or unexpected in relation to the model's existing knowledge about the query—we calculate the informativeness score of the acquired knowledge $k$ given $q$ as follows:
\begin{equation}\small
\begin{split} \small
     S_\texttt{info}(k, q)
     &= 1 - \exp \left(\frac{1}{M} \sum_{i=1}^M \ln P_\theta(k_t|k_{1:t-1}, q)\right)
\end{split}
\end{equation}
\noindent
Assuming the unbiased benchmark model $\theta$ encapsulates world knowledge from general pretraining data, we thus select the GPT-2 series models.

To grasp the expected behaviour of this metric,
consider a simple query: \textit{"What is the capital of the United States?"} The knowledge acquired here is \textit{"Washington"}. In this situation, the model's average probability of generating \textit{"Washington"} is high, as it already knows this fact. Consequently, our informativeness score for this knowledge would be low. Conversely, if the acquired knowledge was \textit{"Chicago"}, the model's probability of generating it would be low. This knowledge is surprising compared to its existing knowledge, resulting in a high informativeness score.
On the other hand, for a tough query where the model is clueless, any provided knowledge would score high on informativeness due to the model's low output probabilities. 
 
\subsection{Extrinsic Evaluation}
Extrinsic evaluation, in contrast to intrinsic evaluation, focuses on uniformly assessing the performance of the acquired knowledge within the context of different downstream tasks. Specifically, we measure how well the acquired knowledge contributes to the downstream task on two types of metrics (\textit{helpfulness} and \textit{validity}). Extrinsic evaluation provides a more comprehensive understanding of the practical value of the acquired knowledge.

\paragraph{Helpfulness} Given a query and answer pair ($q$, $a$), we assess to what extent the acquired knowledge $k$ can help answer the query. As we assume no pre-annotated ground-truth knowledge, 
we use irrelevant knowledge as the baseline. 
Specifically, we randomly sampled $u$ knowledge $\{k^-_{1}, \cdots, k^-_{u}\}$ to reduce the variance of baseline estimation.
Then the helpfulness score is computed as follows:
\begin{equation}\small
\centering
\begin{split}
\small
 &S_\texttt{help}(q, a, k, k^-_{1}, \cdots, k^-_{u}) \\
 &=\max (0, 1 - \frac{\mathcal{L}(q,k,a)}{\frac{1}{u} \sum\nolimits_{i=1}^{u} \mathcal{L}(q,k^-_{i},a)})  \\ 
 &=\max (0, 1 - \frac{\log P(a|q, k)}{\sum\nolimits_{i=1}^{u} \log P(a|q, k^-_{i})})  \label{eq:helpful}
\end{split}
\end{equation} 
where $\mathcal{L}(q,k,a)$ and $\mathcal{L}(q,k^-_{i},a)$ are cross entropy losses of answer generation using $k$ and $k^-_{i}$ respectively. Ideally, the generated knowledge $k$ can provide enough information and reduce the $\mathcal{L}(q,k,a)$ to zero and then the helpfulness score equals one. The worst case is the generated knowledge is no better than random knowledge ($\mathcal{L}(q,k,a) \geq \frac{1}{u} \sum_{i=1}^{u} \mathcal{L}(q,k^-_{i},a)$), and the helpfulness score is naturally zero.

\begin{table*}[!t]
\setlength{\abovecaptionskip}{5pt}   
\setlength{\belowcaptionskip}{0pt}
\setlength{\tabcolsep}{0.92mm}
\small
\centering
\begin{tabular}{ll|ccccccc|cc}
\toprule
 \multirow{2}{*}{\bf Model} & \multirow{2}{*}{\bf Setting} & \multicolumn{3}{c}{\bf Factuality} & \multirow{2}{*}{\bf Relevance} & \multicolumn{2}{c}{\bf Coherence} & \multirow{2}{*}{\bf Inform.} & \multirow{2}{*}{\bf Helpful.} & \multirow{2}{*}{\bf Validity}\\

& & Fact-cons. & Non-verif. & Fact-incon. &  & Coh-sent. & Coh-para. & & &  \\

\midrule
\textsc{DPR} & Supervised & \bf 97.78\% & 2.23\% & 0.00\% & 0.7514 & 0.0301 & 0.7194 & \bf 0.8965 & 0.1236 & 36.86\% \\

\midrule
\textsc{FLAN-T5} & & 58.40\% & 27.80\% & 13.80\% & 0.6848 & \bf \underline{0.1249} & 0.7776 & 0.6727 & 0.0000 & 32.47\%  \\

\textsc{LLaMA} & Zero-shot & \underline{94.20\%} & 4.80\% & 1.00\% & 0.7316 & 0.1183 & 0.8240 & \underline{0.7572} & \underline{0.2191} & 42.00\% \\

\textsc{ChatGPT} &  & 83.63\% & 13.6\% & 2.77\% & \underline{0.8491} & 0.0909 & \bf \underline{0.9033} & 0.7330 & 0.1461 & \bf \underline{43.35\%} \\

\midrule
\textsc{FLAN-T5}  &  & 20.75\% & 62.40\% & 25.40\% & 0.6787 & 0.0416 & 0.8110 & 0.6899 & 0.0000 & 34.65\% \\

\textsc{LLaMA}  & Few-shot & \underline{89.00\%} & 9.20\% & 1.80\% & 0.6966 & \underline{0.0776} & 0.8550 & \underline{0.8545} & \bf \underline{0.2528} & 40.49\% \\

\textsc{ChatGPT}  &  & 86.07\% & 10.97\% & 2.96\% & \bf \underline{0.9205} & 0.0653 & \underline{0.8837} &  0.7700 & 0.1966 & \underline{42.36\%} \\

\bottomrule
\end{tabular}

\caption{\label{tab:nq_result}Automatic evaluation results of different LLMs in the Natural Question test set. \underline{Underlined} and \textbf{Bold} results denote the best results among each setting and among all settings, respectively.}

\vspace{-2mm}
\end{table*}

\begin{table*}[t!]
\setlength{\abovecaptionskip}{5pt}   
\setlength{\belowcaptionskip}{0pt}
\setlength{\tabcolsep}{0.90mm}
\small
\centering
\begin{tabular}{ll|ccccccc|cc}
\toprule
 \multirow{2}{*}{\bf Model} & \multirow{2}{*}{\bf Setting} & \multicolumn{3}{c}{\bf Factuality} & \multirow{2}{*}{\bf Relevance} & \multicolumn{2}{c}{\bf Coherence} & \multirow{2}{*}{\bf Inform.} & \multirow{2}{*}{\bf Helpful.} & \multirow{2}{*}{\bf Validity}\\

& & Fact-cons. & Non-verif. & Fact-incon. &  & Coh-sent. & Coh-para. & & &  \\

\midrule
\textsc{DPR} & Supervised & \bf 91.96\% & 5.18\% & 2.87\% & 0.0907 & 0.0223 & 0.6569 & \bf 0.9357 & 0.0000 & 61.52\% \\

\midrule
\textsc{FLAN-T5} & & 77.90\% & 17.28\% & 4.82\% & 0.3776 & \underline{0.1203} & 0.8331 & 0.7239 & 0.0904 & 56.97\% \\

\textsc{LLaMA} & Zero-shot & \underline{89.46\%} & 8.89\% & 1.65\% & 0.5041 & 0.0548 & 0.8389 & \underline{0.7889} & \bf \underline{0.1178} & \underline{63.50\%} \\

\textsc{ChatGPT} &  & 88.51\% & 10.38\% & 1.11\% & \bf \underline{0.5283} & 0.1028 & \bf \underline{0.9250} & 0.7448 & 0.1023 & 59.76\% \\

\midrule
\textsc{FLAN-T5}  &  & 76.50\% & 17.20\% & 6.30\% & 0.4463 & \bf \underline{0.1523} & 0.7988 & 0.6983 & 0.0934 & 57.18\% \\

\textsc{LLaMA}  & Few-shot & 85.07\% & 12.05\% & 2.88\% & 0.3930 & 0.1088 & 0.7947 & 0.7855 & 0.1132 & \bf \underline{63.79\%} \\

\textsc{ChatGPT}  &  & \underline{85.75\%} & 12.01\% & 2.24\% & \underline{0.4618} & 0.0979 & \underline{0.8632} & \underline{0.7922} & \underline{0.1164} & 60.27\% \\

\bottomrule
\end{tabular}

\caption{\label{tab:wow_result}Automatic evaluation results of different LLMs in the Wizard of Wikipedia test set. }

\vspace{-5mm}
\end{table*}

\paragraph{Validity} To measure how the reliability of the acquired knowledge affects the factuality of the generated answer $a$ on downstream tasks, we define the validity metric for two types of downstream tasks: span-based answers (\textit{e.g.}, open-domain QA) and open-ended answers (\textit{e.g.}, knowledge-grounded dialogue). 
As for span-based answers, the generated answers cannot form a complete sentence for factuality measurement. To this end, we concatenate $(q, a^*)$ as the premise and $(q, a)$ as the hypothesis for deriving the \texttt{factual-consistent} score of the $\mathrm{NLI}(\cdot)$ model as the validity score: 
\begin{equation}\small
S_\texttt{val}(q, a^*, a)= \mathrm{NLI_{fact}}((q, a), (q, a^*))
\end{equation} 
where $a^*$ denotes the ground-truth answer for downstream tasks and the $\mathrm{NLI}(\cdot)$ model is the same as that of  Eq. (\ref{eq:factuality}). 

We demonstrate this measure outperforms traditional metrics like Exact Match and F1 score as shown in Table \ref{tab:bl_col}, which rely on literal matches, and often yield low recall. For instance, an entity pair like 'PRC' and 'China' would receive a zero score due to their differing literal presentations.

As for open-ended answers, we collect $l$ evidence $E=\{e_{1},\dots,e_{l}\}$ and adjust Eq. (\ref{eq:factuality}) to be:
\begin{equation}\small
 S_\texttt{val}(a, E)=f(a, E)=\max_{i=1..l}\mathrm{NLI_{fact}}(a, e_{i})
\end{equation}

\section{Evaluation}
In this section, we will first validate our proposed metrics, and then leverage them to comprehensively evaluate three different types of LLMs across two knowledge-intensive tasks, followed by an in-depth analysis of the results.

\subsection{Metrics Efficacy Validation}
To validate the effectiveness of our proposed metrics, we conducted manual evaluations and compared the results with baseline metrics. Specifically, we developed specific annotation guidelines for each metric, 
detailed in Appendix \ref{sec:detail_human_eval},
and performed manual annotations accordingly. These annotations allowed us to calculate the correlation between each metric and human evaluations. Subsequently, we compared these correlations with baseline metrics (Table \ref{tab:bl_col}). Our metrics demonstrated a strong correlation with human evaluations, significantly outperforming the baseline metrics. 
Details are presented in Chapter \ref{sec:human} and Appendix \ref{sec:detail_human_eval}.

\subsection{Experimental Setups}

\newcommand{\cmark}{\ding{51}}%
\newcommand{\xmark}{\ding{55}}%

\begin{table*}[t!]
\setlength{\abovecaptionskip}{5pt}   
\setlength{\belowcaptionskip}{0pt}
\centering
\small
\begin{adjustbox}{max width=0.95\textwidth}
\begin{tabular}{l|p{2.5cm}|p{8.7cm}|l}
\toprule
Model & Query & Acquired Knowledge & Answer \\
\midrule
LLaMA & Who wrote the song the glory of love? & "The Glory of Love" is a song written by Billy Hill, recorded by American \textcolor{blue}{singer} Benny Goodman in 1936. The song has been covered by many other artists, including a version by American singer Peggy Lee with \textcolor{blue}{Bing Crosby and the Johnny Mann} Singers in 1959. & Billy Hill \checkmark \\

\midrule
FLAN-T5 & Who wrote the song the glory of love? & The Glory of Love is a song written by the American composer and lyricist \textcolor{red}{Irving Berlin}. & Irving Berlin \xmark \\

\midrule
DPR & Rain sleet or snow that contains a high concentration of acids is called? & ... Ice pellets are a form of precipitation consisting of small, translucent balls of ice. This form of precipitation is also referred to as ""sleet"" by the United States National Weather Service. (In British English ""sleet"" refers to a mixture of rain and snow)... & icy rain \xmark \\

\bottomrule
\end{tabular}
\end{adjustbox}
\caption{Factuality of acquired knowledge may not influence the validity of the answer. \textcolor{red}{Red} words represent factual errors in critical information, while \textcolor{blue}{blue} words represent factual errors in non-critical information.}
\vspace{-0.1in}
\vspace{0.2in}
\label{tab:factulity_relevance}
\vspace{-5mm}
\end{table*}




 



\begin{table}[t!]
\setlength{\abovecaptionskip}{5pt}   
\setlength{\belowcaptionskip}{0pt}
\setlength{\tabcolsep}{0.54mm}
\small
\centering
\begin{tabular}{llccccc}
\toprule
\multirow{2}{*}{\bf Model} & \multirow{2}{*}{\bf Extrinsic} & \multicolumn{5}{c}{\bf Instrinsic} \\
\cmidrule(lr){3-7}
& & Fact. & Rel. & Coh-{sent.} & Coh-{para.} & Info. \\

\midrule
\multirow{2}{*}{\textsc{DPR}} & helpful. & 0.10 & \textbf{0.24}$^\dagger$ & 0.07 & -0.03 & -0.14$^\dagger$ \\
 & validity & 0.04 & \textbf{0.19}$^\dagger$ & 0.04 & -0.06 & -0.09 \\

\midrule
\multirow{2}{*}{\textsc{LLMs}} & helpful. & \textbf{0.14} & -0.05 & 0.10 & -0.09 & -0.05 \\
 & validity & \textbf{0.15}$^\dagger$ & -0.02 & 0.07 & -0.03 & -0.03 \\
 
\bottomrule
\vspace{-2mm}
\end{tabular}

\caption{\label{tab:inter_correlation_nq} The Somers' correlation between intrinsic and extrinsic metrics on NQ. Scores with $p\text{-}value < 0.05$ are marked with $^\dagger$. \textbf{Bold} results denote the most correlated intrinsic metric to the concerned extrinsic metric. The breakdowns of all correlations are in Appendix~\ref{sec:break_down}.}
\vspace{-1mm}
\end{table}

\begin{table}[t!]
\setlength{\abovecaptionskip}{5pt}   
\setlength{\belowcaptionskip}{0pt}
\setlength{\tabcolsep}{0.54mm}
\small
\centering
\begin{tabular}{llccccc}
\toprule
\multirow{2}{*}{\bf Model} & \multirow{2}{*}{\bf Extrinsic} & \multicolumn{5}{c}{\bf Instrinsic} \\

\cmidrule(lr){3-7}
& & Fact. & Rel. & Coh-{sent.} & Coh-{para.} & Info. \\

\midrule
\multirow{2}{*}{\textsc{DPR}} & helpful. & 0.01 & \textbf{0.27}$^\dagger$ & 0.10$^\dagger$ & -0.03 & -0.14$^\dagger$ \\
 & validity & -0.01 & -0.06 & \textbf{0.13}$^\dagger$ & -0.12$^\dagger$ & -0.13$^\dagger$ \\

\midrule
\multirow{2}{*}{\textsc{LLMs}} & helpful. & 0.06 & 0.05 & \textbf{0.10} & 0.00 & -0.16 \\
 & validity & \textbf{0.24}$^\dagger$ & 0.09 & 0.05 & -0.02 & -0.07 \\
 
\bottomrule
\end{tabular}

\caption{\label{tab:inter_correlation} The Somers' correlation between intrinsic and extrinsic metrics on WoW. }
\vspace{-5mm}
\end{table}

\paragraph{Baselines}
Compared with a popular retrieval-based model, DPR \cite{karpukhin2020dense}, we evaluate knowledge generation with the three different types of LLMs, including FLAN-T5 \cite{wei2022finetuned}, LLaMA \cite{touvron2023llama}, and ChatGPT \cite{ouyang2022training}. 
By default, we report the results with the largest size of each LLM and adopt greedy decoding in our experiments for reproducibility. Details are presented in Appendix~\ref{app:baseline}.  

\vspace{-1mm}

\paragraph{Datasets}
We evaluate the generated knowledge on two widely-studied benchmark datasets, including 1) \textbf{Natural Questions (NQ)} \cite{kwiatkowski-etal-2019-natural}, an open-domain QA dataset; and 2) \textbf{Wizard of Wikipedia (WoW)} \cite{dinan2018wizard} a knowledge-grounded dialogue dataset. 
During experiments, we randomly sample 500 examples from the NQ and WoW test sets respectively for evaluation. 
Details are presented in Appendix~\ref{app:dataset}.  

\vspace{-1mm}

\paragraph{Implementation Details} All the adopted models in \texttt{CONNER} are introduced in Appendix~\ref{app:implement}. 

\vspace{-1mm}

\paragraph{Evaluation Setting}
Following \cite{yu2023generate}, we evaluate the knowledge generation of LLMs under both zero-shot and few-shot settings. After the knowledge acquisition, we perform QA or dialogue generation under the few-shot setting to further investigate the impact of different knowledge acquisition methods on downstream tasks.

\textit{1) Zero-shot Evaluation}: We test with varied prompts and report peak performance. A prompt could be ``Generate Wikipedia knowledge for the query. \{query\}''. Prompts tried are in Appendix~\ref{sec:knowledge-gen-prompt}.

\textit{2) Few-shot Evaluation}: We construct the prompt with $k$ randomly chosen samples from the training set. The example templates used for knowledge generation are listed in Appendix \ref{sec:knowledge-gen-prompt} and \ref{sec:answer-gen-prompt}.

\subsection{Overall Evaluation}\label{sec:eval}
Table~\ref{tab:nq_result} and Table~\ref{tab:wow_result} summarize the evaluation results of DPR and three LLM-based knowledge generators on NQ and WoW datasets, respectively. There are several notable observations as follows:\vspace{-0.1cm}

\paragraph{Generated knowledge exceeds retrieved knowledge in most evaluation perspectives, except the \textit{factuality} and \textit{informativeness}.} 
In both NQ and WoW scenarios, LLMs show remarkable capabilities in generating highly \textit{relevant} and \textit{coherent} knowledge. Moreover, the knowledge generated by LLMs also proves to be more beneficial for downstream tasks, regarding both \textit{helpfulness} and \textit{validity}. These results highlight the significant advantages of utilizing LLMs as knowledge generators in terms of knowledge quality and applicability, rendering them a valuable knowledge resource for various knowledge-intensive applications. \vspace{-0.1cm}

\paragraph{Despite obtaining lower \textit{factuality} than retrieved knowledge, generated knowledge contributes more to the factuality of downstream tasks (\textit{i.e.}, higher \textit{validity}).} 
To investigate the underlying reason, we analyze the correlation between different intrinsic metrics and extrinsic metrics on two tasks. 
As shown in Tables~\ref{tab:inter_correlation_nq} and~\ref{tab:inter_correlation}, the performance of downstream tasks is indeed hindered by the issue of \textit{factuality} in the generated knowledge from LLMs. However, for retrieval models (\textit{e.g.}, DPR), limitations may arise from the \textit{relevance} and \textit{coherence} of the retrieved knowledge, while its high factuality fails to ensure the performance of downstream tasks. 
We present a case study in Table~\ref{tab:factulity_relevance}, which intuitively shows that the presence of factual errors in non-critical information has minimal impact on downstream tasks, while it is highly impossible to derive the correct answer from the irrelevant retrieved knowledge.   
While LLaMA and ChatGPT generate knowledge with slightly lower factuality than DPR, it is shown to be adequate for downstream tasks.
At this point, the relevance of the acquired knowledge is more critical. 
Hence, relying solely on the factuality of the knowledge itself is an unreliable means of assessing its impact on the factuality of downstream tasks. 
Motivated by this finding, we investigate approaches to guiding the generated knowledge selection with the multi-perspective evaluation outcome of \texttt{CONNER} for improving the downstream performance in $\S$ \ref{improvement}. \vspace{-0.1cm}

\paragraph{DPR falls short of retrieving relevant and helpful knowledge for knowledge-grounded dialogues.} 
As the DPR model is finetuned on QA datasets to match a question to Wikipedia knowledge, the DPR model struggles to match dialogue utterances with the necessary knowledge.
Also, the candidate Wikipedia passages in DPR (100 tokens) are much longer than the knowledge needed in WoW, containing much redundant information. 
This reveals the shortcomings of supervised dense retrieval models, such as limited transferability and being constrained by knowledge bases. \vspace{-0.1cm}

\paragraph{Few-shot in-context learning for LLMs generally harms the \textit{factuality} of generated knowledge.} 
We observe that the length of knowledge generated by few-shot ICL is generally longer than that of zero-shot prompting since the ground-truth knowledge for demonstrations is relatively long. 
Consequently, LLM is more error-prone (see the analysis of \textbf{long-form generation} in $\S$ \ref{length_ana}). 
This indicates that few-shot ICL is not always better than zero-shot ICL in knowledge generation, and the selected demonstrations attach great importance. 
Inspired by this, we investigate approaches to guiding the few-shot demonstration selection with the evaluation outcome of \texttt{CONNER} for improving the performance of few-shot ICL in $\S$ \ref{improvement}.\vspace{-0.1cm}

\paragraph{FLAN-T5 fails to be a qualified knowledge generator since its generated knowledge is \textit{poorly factual} and \textit{rarely helpful} to downstream tasks.} 
Although FLAN-T5 (11B) significantly surpasses many models of the same scale through instruction tuning on numerous tasks, it falls short of being a qualified knowledge generator. As shown in Table~\ref{tab:factulity_relevance}, such a low factuality leads to frequent occurrences of factual errors in critical information, thereby harming downstream tasks. \vspace{-0.1cm}
To this end, we study the scaling of performance w.r.t different perspectives by varying the \textbf{model size} in $\S$ \ref{sec:analysis}.

\subsection{Further Analysis}\label{sec:analysis}

We further analyze how different factors affect the quality and reliability of the generated knowledge and discuss our findings below.\vspace{-0.1cm}

\begin{figure}
\setlength{\abovecaptionskip}{5pt}   
\setlength{\belowcaptionskip}{0pt}
    \centering
    \small
     \includegraphics[width=\linewidth]{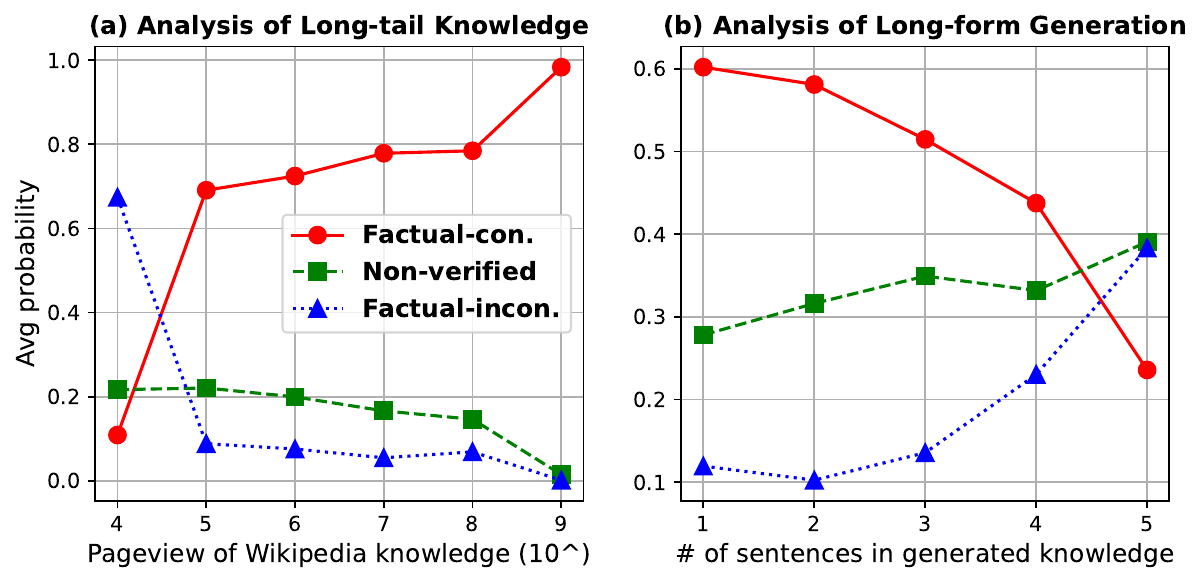}
    \caption{The impact of knowledge frequency and length on the factuality of the generated knowledge.}
    \label{fig:length_ana}
    \vspace{-0.4cm}
\end{figure}

\paragraph{Long-tail Knowledge}
We investigate the impact of the knowledge frequency on the factuality performance of LLaMA on the WoW dataset. Each data entry in WoW comprises a topic, query, knowledge, and answer. The topic indicates the corresponding Wikipedia page linked to the knowledge. We assess this knowledge's frequency using Wikipedia pageviews from 2015 to 2021\footnote{\url{https://wikimedia.org/api/rest_v1}}. This enables us to differentiate between common and long-tail knowledge in WoW. Our findings reveal that LLaMA exhibits lower reliability when it is expected to generate rare/long-tail knowledge compared to common knowledge, as depicted in Figure \ref{fig:length_ana}(a).\vspace{-0.1cm}

\paragraph{Long-form Generation}
\label{length_ana}
We investigate the impact of generation length on the factuality of the generated knowledge. Specifically, we consider knowledge over 40 tokens and take sentences as evaluation units aligned with factuality evaluation.
Figure~\ref{fig:length_ana}(b) displays the factuality performance based on the number of sentences in the generated knowledge. 
The results show that LLaMA exhibits higher error rates when generating long-form knowledge. 
Therefore, prompting the LLMs to generate the required knowledge in a concise rather than lengthy manner can benefit factuality.\vspace{-0.1cm}

\paragraph{Impact of Model Size}
\begin{figure} 
\setlength{\abovecaptionskip}{5pt}   
\setlength{\belowcaptionskip}{0pt}
    \centering
     \includegraphics[width=\linewidth]{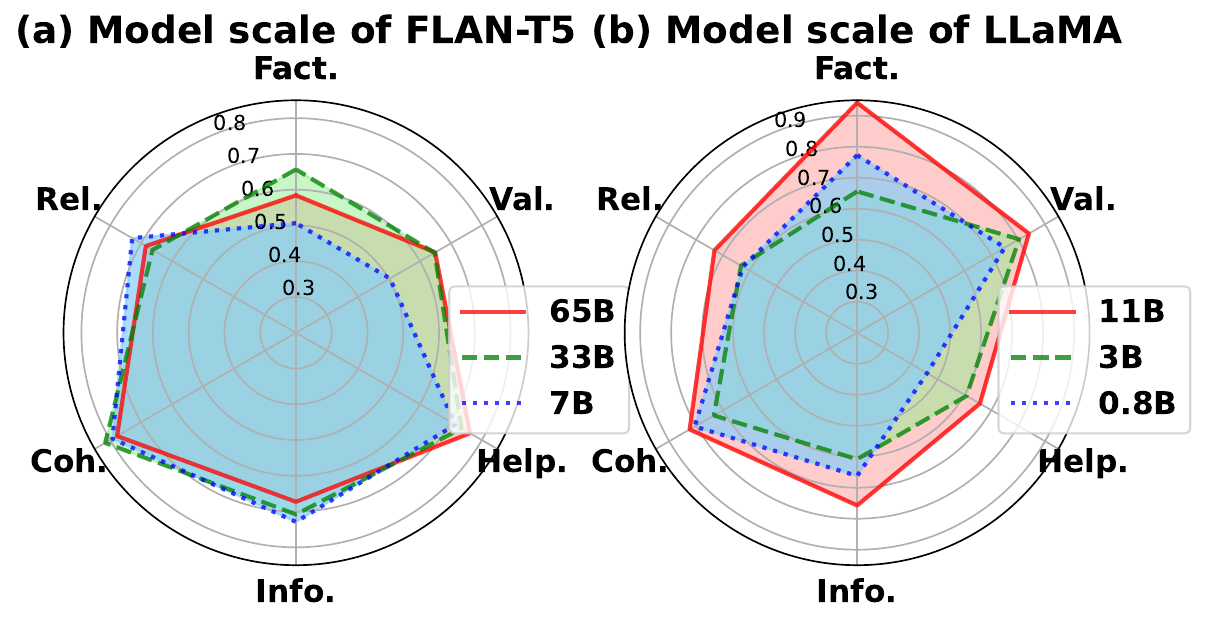}
    \caption{Performance on NQ with different sizes of FLAN-T5 and LLaMA as the knowledge generator (Help. and Val. scores are linearly scaled).}
    \label{model_size_radars}
    \vspace{-0.4cm}
\end{figure}

Figures \ref{model_size_radars} depicts the performance scaling with the model size, including \texttt{LLaMA-65B/33B/7B} and \texttt{FLAN-T5-11B/3B/780M}. The results are reported on the NQ dataset using zero-shot prompting. We observe that larger models do not necessarily outperform smaller models in terms of intrinsic evaluation (particularly when parameter magnitudes are similar). However, larger models consistently outperform smaller models in terms of extrinsic evaluation(helpfulness and validity). Detailed tables are presented in  Appendix~\ref{sec:model_size}.\vspace{-0.1cm}

\section{Two Use Cases of \texttt{CONNER}}
\label{improvement}

To explore how our framework can guide the future design of utilizing LLMs as a knowledge generator, we design two strategies to employ \texttt{CONNER} as a measurement for guiding the \textbf{Prompt Engineering} and \textbf{Knowledge Selection} for knowledge-intensive tasks.
We define the overall quality of knowledge $k$ given the query $q$ as follows: 
\begin{equation} \small
\begin{split}
  &Q_\texttt{know}(q, k) = \mathbf{\gamma}^\intercal \cdot \mathbf{S}_\texttt{intr} \quad \mathbf{\gamma} \in \mathbb{R}^4  \\
 &\mathbf{S_\texttt{intr}}=[S_\texttt{fact},S_\texttt{rel},S_\texttt{coh\_para},S_\texttt{info}]^\intercal 
\end{split}
\end{equation} 
where $Q_\texttt{know}$ is the linear combination of four instinct metrics $\mathbf{S_\texttt{intr}}$ and $\mathbf{\gamma}$ is the coefficient vector.

\paragraph{Prompt Engineering}
We show how to use \texttt{CONNER} to improve knowledge generation by performing prompt engineering for few-shot ICL. 
We random sample a small set of $m$ samples from the training set, then use $Q_\texttt{know}(q, k)$ as the scoring function to select the top $n$ samples to compose the few-shot prompt. As shown in Table~\ref{tab:prompt_eng}, the knowledge generated by \texttt{CONNER}-enhanced few-shot prompting outperforms that with random demonstrations on 3 out of 4 perspectives, under the setting of $m=30$ and $n=8$. 

\begin{table}[t!]
\setlength{\abovecaptionskip}{5pt}   
\setlength{\belowcaptionskip}{0pt}
\setlength{\tabcolsep}{1.45mm}
\small
\centering
\begin{tabular}{l|cccc}
\toprule
\textbf{Model} & Fact. & Rel. & Coh. & Info. \\

\midrule
ChatGPT  & 85.8\% & 0.462 & 0.863 & \bf 0.792 \\
ChatGPT$_{select\ prompt}$ & \bf 87.7\% & \bf 0.503 & \bf 0.899 & 0.775 \\

\bottomrule
\end{tabular}

\caption{\label{tab:prompt_eng} \texttt{CONNER}-guided demonstration selection improves the intrinsic quality of generated knowledge.}
\vspace{-1mm}
\end{table}

\paragraph{Knowledge Selection}
We employ \texttt{CONNER} to improve downstream tasks by selecting high-quality generated knowledge. Specifically, we generate $r$ different knowledge $\mathcal{H}=\{\tilde{k_1},...,\tilde{k_r}\}$ from LLMs with top-$p$ sampling, then select the generated knowledge for the downstream task, according to  
$k = \argmax\nolimits_{\tilde{k} \in \mathcal{H}} Q_\texttt{know}(q, \tilde{k})$. 
As shown in Table~\ref{tab:ranking}, 
we achieve a relative improvement of 43.15\% in helpfulness on ChatGPT with $p=0.9$ and $r=5$. 

\begin{table}[t!]
\setlength{\abovecaptionskip}{5pt}   
\setlength{\belowcaptionskip}{0pt}
\setlength{\tabcolsep}{2.25mm}
\small
\centering
\begin{tabular}{l|cc}
\toprule
\textbf{Model} & Helpfulness & Validity  \\

\midrule
ChatGPT & 0.1461 & 43.45\% \\
ChatGPT$_{select\ knowledge}$ & \bf 0.2090 & \bf 44.28\% \\

\bottomrule
\end{tabular}
\caption{\label{tab:ranking} \texttt{CONNER}-guided knowledge selection improves extrinsic (downstream) performance.}
\vspace{-4mm}
\end{table}

\section{Human Evaluation}\label{sec:human}

We conducted a human evaluation by randomly selecting 400 samples from the NQ and WoW test sets. Our three annotators provided ratings for the intrinsic and extrinsic metrics for the four models. Additionally, for FLAN-T5 and LLaMA, we annotated the specific locations of factual errors in the generated knowledge, aiming to facilitate future research on fine-grained fallacy detection. Detailed annotation instructions and the statistics of our labelled data can be found in Appendix~\ref{sec:label_instruct}.

To evaluate how well \texttt{CONNER} matches human evaluation of knowledge and compares with several baseline metrics, 
we measure the Somers’ D correlation~\cite{somers1962new} between the human rating {0, 1, 2} of the knowledge quality and corresponding metric scores. 
Table~\ref{tab:human_col} and Table~\ref{tab:bl_col} illustrate the results of four models on the NQ dataset. 
We observe that: (1) \texttt{CONNER} yields consistently good correlations with human evaluation w.r.t different evaluation perspectives (except for informativeness), which indicates that the quality of knowledge can be more effectively evaluated with \texttt{CONNER}.  
The inconsistency between informativeness and human judgment is attributed to the differences in model knowledge and human knowledge. 
(2) \texttt{CONNER}  metrics consistently outperform all other reference-reliant metrics,
indicating the effectiveness of our framework in the knowledge evaluation scenarios.






\begin{table}[t!]
\setlength{\abovecaptionskip}{5pt}   
\setlength{\belowcaptionskip}{0pt}
\setlength{\tabcolsep}{1.25mm}
\small
\centering
\begin{tabular}{l|c|c|c|c}
\toprule

Metric & DPR & FLAN-T5 & LLaMA & ChatGPT \\
\midrule

Factuality & 0.65$^\dagger$ & 0.66$^\dagger$ & 0.66$^\dagger$ & 0.63$^\dagger$ \\
\midrule
Relevance & 0.69$^\dagger$ & 0.37$^\dagger$ & 0.55$^\dagger$ & 0.54$^\dagger$ \\
\midrule
Coherence & 0.53$^\dagger$ & 0.58$^\dagger$ & 0.44$^\dagger$ & 0.49$^\dagger$ \\
\midrule
Informative & 0.30$^\dagger$ & 0.17 & 0.35 & 0.32$^\dagger$ \\
\midrule
Helpfulness & 0.75$^\dagger$ & 0.45$^\dagger$ & 0.81$^\dagger$ & 0.69$^\dagger$ \\
\midrule
Validity & 0.83$^\dagger$ & 0.73$^\dagger$ & 0.85$^\dagger$ & 0.82$^\dagger$ \\

\bottomrule
\end{tabular}

\caption{\label{tab:human_col} Somer's D correlation of metrics with the human annotation on NQ (The results on WoW are presented in Appendix~\ref{app:wow_human}).
Correlation scores with $p\text{-}value < 0.05$ are marked with $^\dagger$.
}

\vspace{-2mm}
\end{table}






\begin{table}[t!]
\setlength{\abovecaptionskip}{5pt}   
\setlength{\belowcaptionskip}{0pt}
\setlength{\tabcolsep}{1.25mm}
\small
\centering
\begin{tabular}{l|c|c|c|c}
\toprule

Metric & DPR & FLAN-T5 & LLaMA & ChatGPT \\
\midrule

Factuality & 0.65$^\dagger$ & 0.66$^\dagger$ & 0.66$^\dagger$ & 0.63$^\dagger$ \\

HE & -0.24 & 0.15 & -0.03 & 0.29$^\dagger$  \\

NLI & 0.23 & 0.47$^\dagger$  & 0.27$^\dagger$  & 0.38$^\dagger$  \\

NLI-Multitask & 0.18$^\dagger$  & 0.51$^\dagger$ & 0.26$^\dagger$  & 0.32$^\dagger$ \\

NLI-Decompose. & 0.23$^\dagger$  & 0.47$^\dagger$  & 0.27$^\dagger$ & 0.38$^\dagger$ \\

\midrule
Relevance & 0.69$^\dagger$ & 0.37$^\dagger$ & 0.55$^\dagger$ & 0.54$^\dagger$ \\

F1 & 0.45$^\dagger$  & 0.21 & 0.41$^\dagger$  & 0.47$^\dagger$  \\

\midrule
Validity & 0.83$^\dagger$ & 0.73$^\dagger$ & 0.85$^\dagger$ & 0.82$^\dagger$ \\

EM & 0.59$^\dagger$  & 0.51$^\dagger$ & 0.54$^\dagger$  & 0.61$^\dagger$  \\

F1 & 0.74$^\dagger$  & 0.67$^\dagger$ & 0.76$^\dagger$  & 0.77$^\dagger$  \\ 



\bottomrule
\end{tabular}

\caption{\label{tab:bl_col} Comparing \texttt{CONNER} with reference-reliant baseline metrics on the NQ dataset. Details of baseline metrics are presented in Appendix \ref{sec:bl_metrics}. 
}
\vspace{-4mm}
\end{table}

\section{Conclusion}
In this work, we introduce \texttt{CONNER}, a comprehensive evaluation framework designed to automatically assess both the intrinsic quality and extrinsic reliability of the knowledge generated by LLMs. 
Notably, \texttt{CONNER} is reference-free but demonstrates a better correlation with human judgement compared with previous reference-reliant metrics.

Through extensive evaluation and in-depth analysis, we identify several key factors affecting the factuality of generated knowledge. We find
although the generated knowledge is less factual than the retrieved knowledge, it remarkably enhances the factuality of downstream tasks over the retrieved ones. Furthermore, we propose two approaches to improve knowledge generation and downstream task performance with the guidance of \texttt{CONNER}. 
We believe our framework and findings will facilitate the future research of trustworthy AIGC.

\section*{Limitations}

In this section, we discuss the limitations in this work from three perspectives. 

Firstly, the knowledge we evaluate primarily relies on information sourced from Wikipedia. This choice is driven by two considerations: (1) Large language models (LLMs) are trained on diverse corpora, which may include undisclosed domain-specific or task-specific data. To ensure fairness in our evaluations and enable meaningful comparisons, we focus on the common data sources that all models have learned from, with Wikipedia being a prevalent pre-training corpus for different LLMs. (2) Wikipedia is renowned for its high-quality knowledge, providing us with authoritative evidence to validate the generated knowledge. Additionally, leveraging such authoritative evidence enhances the interpretability of our factual judgments. In future work, we aim to expand our evaluations to include a broader range of world knowledge, thus further enhancing the scope and generalizability of our findings.

Secondly, while our work primarily aims to propose a general framework that can be applied to any language, our evaluation framework presents potential generalization challenges for non-English languages. This is due to its reliance on several common NLP components, a limitation echoed across many NLP methodologies. Encouragingly, the development of model variants in other languages, such as Chinese \cite{hu-etal-2020-ocnli,xie2023t2ranking,chi_discourse}, indicates the potential for broader applications. Nonetheless, the reality remains that for very low-resource languages without existing NLP models, these components may need to be developed from scratch. This issue represents a challenge that the community needs to address in the future.

A third limitation is that our assessment of factuality is limited to sentence-level granularity. Through analysis and manual annotation, we have identified that large language models (LLMs) tend to exhibit errors at a more detailed level, particularly concerning numbers, time, and the generation of misleading or fabricated concepts (e.g., key characters, identities, and locations), particularly within parallel structures. To address this limitation, future research will concentrate on developing more fine-grained methods for detecting hallucinations and assessing factual accuracy. To facilitate such research, we have annotated a specific subset of data that targets fine-grained factual errors.

Despite these limitations, we believe our work serves as a significant catalyst for the automated evaluation of knowledge generated by large language models, contributing positively to the advancement of more trustworthy AI systems.

\section*{Acknowledgements}
We extend our sincerest gratitude to Professor Jing Ma, whose insightful discussions and suggestions on factuality evaluation have significantly inspired our design. We are particularly grateful to our three anonymous reviewers, whose thorough and meticulous reviews have considerably improved the quality of our work. Their constructive discussions and insights have undoubtedly enhanced our revisions. 
This research work is partially supported by CUHK under Project No. 3230377 (Ref. No. KPF23GW20).

\bibliography{emnlp2023}

\begin{thebibliography}{56}
\expandafter\ifx\csname natexlab\endcsname\relax\def\natexlab#1{#1}\fi

\bibitem[{Black et~al.(2021)Black, Gao, Wang, Leahy, and Biderman}]{gpt-neo}
Sid Black, Leo Gao, Phil Wang, Connor Leahy, and Stella Biderman. 2021.
\newblock \href {https://doi.org/10.5281/zenodo.5297715} {{GPT-Neo}: Large scale autoregressive language modeling with {Mesh-Tensorflow}}.

\bibitem[{Braun and Clarke(2012)}]{theme_ana}
Virginia Braun and Victoria Clarke. 2012.
\newblock \emph{Thematic analysis.}, pages 57--71.

\bibitem[{Chen et~al.(2023)Chen, Wang, Deng, Kwan, Wang, and Wong}]{chen-etal-2023-towards-robust}
Liang Chen, Hongru Wang, Yang Deng, Wai~Chung Kwan, Zezhong Wang, and Kam-Fai Wong. 2023.
\newblock \href {https://doi.org/10.18653/v1/2023.findings-acl.462} {Towards robust personalized dialogue generation via order-insensitive representation regularization}.
\newblock In \emph{Findings of the Association for Computational Linguistics: ACL 2023}, pages 7337--7345, Toronto, Canada. Association for Computational Linguistics.

\bibitem[{Deng et~al.(2023{\natexlab{a}})Deng, Lei, Huang, and Chua}]{acl23-tutorial}
Yang Deng, Wenqiang Lei, Minlie Huang, and Tat{-}Seng Chua. 2023{\natexlab{a}}.
\newblock \href {https://doi.org/10.18653/v1/2023.acl-tutorials.1} {Goal awareness for conversational {AI:} proactivity, non-collaborativity, and beyond}.
\newblock In \emph{Proceedings of the 61st Annual Meeting of the Association for Computational Linguistics: Tutorial Abstracts, {ACL} 2023, Toronto, Canada, July 9-14, 2023}, pages 1--10. Association for Computational Linguistics.

\bibitem[{Deng et~al.(2023{\natexlab{b}})Deng, Zhang, Yuan, and Lam}]{acl23-esc}
Yang Deng, Wenxuan Zhang, Yifei Yuan, and Wai Lam. 2023{\natexlab{b}}.
\newblock \href {https://doi.org/10.18653/v1/2023.acl-long.225} {Knowledge-enhanced mixed-initiative dialogue system for emotional support conversations}.
\newblock In \emph{Proceedings of the 61st Annual Meeting of the Association for Computational Linguistics (Volume 1: Long Papers), {ACL} 2023}, pages 4079--4095. Association for Computational Linguistics.

\bibitem[{Dinan et~al.(2018)Dinan, Roller, Shuster, Fan, Auli, and Weston}]{dinan2018wizard}
Emily Dinan, Stephen Roller, Kurt Shuster, Angela Fan, Michael Auli, and Jason Weston. 2018.
\newblock Wizard of wikipedia: Knowledge-powered conversational agents.
\newblock \emph{arXiv preprint arXiv:1811.01241}.

\bibitem[{Dziri et~al.(2022)Dziri, Kamalloo, Milton, Zaiane, Yu, Ponti, and Reddy}]{dziri2022faithdial}
Nouha Dziri, Ehsan Kamalloo, Sivan Milton, Osmar Zaiane, Mo~Yu, Edoardo~M. Ponti, and Siva Reddy. 2022.
\newblock \href {http://arxiv.org/abs/2204.10757} {Faithdial: A faithful benchmark for information-seeking dialogue}.

\bibitem[{Glover et~al.(2022{\natexlab{a}})Glover, Fancellu, Jagannathan, Gormley, and Schaaf}]{revisit_nli}
John Glover, Federico Fancellu, Vasudevan Jagannathan, Matthew~R. Gormley, and Thomas Schaaf. 2022{\natexlab{a}}.
\newblock \href {https://doi.org/10.48550/arXiv.2211.16853} {Revisiting text decomposition methods for nli-based factuality scoring of summaries}.
\newblock \emph{CoRR}, abs/2211.16853.

\bibitem[{Glover et~al.(2022{\natexlab{b}})Glover, Fancellu, Jagannathan, Gormley, and Schaaf}]{glover2022revisiting}
John Glover, Federico Fancellu, Vasudevan Jagannathan, Matthew~R. Gormley, and Thomas Schaaf. 2022{\natexlab{b}}.
\newblock \href {http://arxiv.org/abs/2211.16853} {Revisiting text decomposition methods for nli-based factuality scoring of summaries}.

\bibitem[{Honovich et~al.(2021)Honovich, Choshen, Aharoni, Neeman, Szpektor, and Abend}]{Q2}
Or~Honovich, Leshem Choshen, Roee Aharoni, Ella Neeman, Idan Szpektor, and Omri Abend. 2021.
\newblock \href {http://arxiv.org/abs/2104.08202} {$q^{2}$: Evaluating factual consistency in knowledge-grounded dialogues via question generation and question answering}.

\bibitem[{Hu et~al.(2020)Hu, Richardson, Xu, Li, K{\"u}bler, and Moss}]{hu-etal-2020-ocnli}
Hai Hu, Kyle Richardson, Liang Xu, Lu~Li, Sandra K{\"u}bler, and Lawrence Moss. 2020.
\newblock \href {https://doi.org/10.18653/v1/2020.findings-emnlp.314} {{OCNLI}: {O}riginal {C}hinese {N}atural {L}anguage {I}nference}.
\newblock In \emph{Findings of the Association for Computational Linguistics: EMNLP 2020}, pages 3512--3526, Online. Association for Computational Linguistics.

\bibitem[{Huang et~al.(2017)Huang, Tan, Huang, Mo, and Zhou}]{chi_discourse}
Guimin Huang, Min Tan, Sirui Huang, Ruyu Mo, and Ya~Zhou. 2017.
\newblock \href {https://doi.org/10.1109/PIC.2017.8359586} {A discourse coherence model for analyzing chinese students' essay}.
\newblock In \emph{2017 International Conference on Progress in Informatics and Computing (PIC)}, pages 430--434.

\bibitem[{Izacard and Grave(2021)}]{izacard2021leveraging}
Gautier Izacard and Edouard Grave. 2021.
\newblock Leveraging passage retrieval with generative models for open domain question answering.
\newblock In \emph{EACL 2021}, pages 874--880.

\bibitem[{Jagerman et~al.(2023)Jagerman, Zhuang, Qin, Wang, and Bendersky}]{query_expansion}
Rolf Jagerman, Honglei Zhuang, Zhen Qin, Xuanhui Wang, and Michael Bendersky. 2023.
\newblock \href {https://doi.org/10.48550/arXiv.2305.03653} {Query expansion by prompting large language models}.
\newblock \emph{CoRR}, abs/2305.03653.

\bibitem[{Ji et~al.(2023)Ji, Lee, Frieske, Yu, Su, Xu, Ishii, Bang, Madotto, and Fung}]{hallucination_survey_2023}
Ziwei Ji, Nayeon Lee, Rita Frieske, Tiezheng Yu, Dan Su, Yan Xu, Etsuko Ishii, Ye~Jin Bang, Andrea Madotto, and Pascale Fung. 2023.
\newblock \href {https://doi.org/10.1145/3571730} {Survey of hallucination in natural language generation}.
\newblock \emph{ACM Comput. Surv.}, 55(12).

\bibitem[{Jwalapuram et~al.(2021)Jwalapuram, Joty, and Lin}]{para_coh}
Prathyusha Jwalapuram, Shafiq~R. Joty, and Xiang Lin. 2021.
\newblock \href {http://arxiv.org/abs/2110.07198} {Rethinking self-supervision objectives for generalizable coherence modeling}.
\newblock \emph{CoRR}, abs/2110.07198.

\bibitem[{Kadavath et~al.(2022{\natexlab{a}})Kadavath, Conerly, Askell, Henighan, Drain, Perez, Schiefer, Dodds, DasSarma, Tran-Johnson et~al.}]{kadavath2022language}
Saurav Kadavath, Tom Conerly, Amanda Askell, Tom Henighan, Dawn Drain, Ethan Perez, Nicholas Schiefer, Zac~Hatfield Dodds, Nova DasSarma, Eli Tran-Johnson, et~al. 2022{\natexlab{a}}.
\newblock Language models (mostly) know what they know.
\newblock \emph{arXiv preprint arXiv:2207.05221}.

\bibitem[{Kadavath et~al.(2022{\natexlab{b}})Kadavath, Conerly, Askell, Henighan, Drain, Perez, Schiefer, Hatfield{-}Dodds, DasSarma, Tran{-}Johnson, Johnston, Showk, Jones, Elhage, Hume, Chen, Bai, Bowman, Fort, Ganguli, Hernandez, Jacobson, Kernion, Kravec, Lovitt, Ndousse, Olsson, Ringer, Amodei, Brown, Clark, Joseph, Mann, McCandlish, Olah, and Kaplan}]{LM_know}
Saurav Kadavath, Tom Conerly, Amanda Askell, Tom Henighan, Dawn Drain, Ethan Perez, Nicholas Schiefer, Zac Hatfield{-}Dodds, Nova DasSarma, Eli Tran{-}Johnson, Scott Johnston, Sheer~El Showk, Andy Jones, Nelson Elhage, Tristan Hume, Anna Chen, Yuntao Bai, Sam Bowman, Stanislav Fort, Deep Ganguli, Danny Hernandez, Josh Jacobson, Jackson Kernion, Shauna Kravec, Liane Lovitt, Kamal Ndousse, Catherine Olsson, Sam Ringer, Dario Amodei, Tom Brown, Jack Clark, Nicholas Joseph, Ben Mann, Sam McCandlish, Chris Olah, and Jared Kaplan. 2022{\natexlab{b}}.
\newblock \href {https://doi.org/10.48550/arXiv.2207.05221} {Language models (mostly) know what they know}.
\newblock \emph{CoRR}, abs/2207.05221.

\bibitem[{Karpukhin et~al.(2020)Karpukhin, Oguz, Min, Lewis, Wu, Edunov, Chen, and Yih}]{karpukhin2020dense}
Vladimir Karpukhin, Barlas Oguz, Sewon Min, Patrick Lewis, Ledell Wu, Sergey Edunov, Danqi Chen, and Wen-tau Yih. 2020.
\newblock Dense passage retrieval for open-domain question answering.
\newblock In \emph{Proceedings of the 2020 Conference on Empirical Methods in Natural Language Processing (EMNLP)}, pages 6769--6781.

\bibitem[{Komeili et~al.(2021)Komeili, Shuster, and Weston}]{komeili2021internetaugmented}
Mojtaba Komeili, Kurt Shuster, and Jason Weston. 2021.
\newblock \href {http://arxiv.org/abs/2107.07566} {Internet-augmented dialogue generation}.

\bibitem[{Kryscinski et~al.(2020{\natexlab{a}})Kryscinski, McCann, Xiong, and Socher}]{factcc}
Wojciech Kryscinski, Bryan McCann, Caiming Xiong, and Richard Socher. 2020{\natexlab{a}}.
\newblock \href {https://doi.org/10.18653/v1/2020.emnlp-main.750} {Evaluating the factual consistency of abstractive text summarization}.
\newblock In \emph{Proceedings of the 2020 Conference on Empirical Methods in Natural Language Processing (EMNLP)}, pages 9332--9346, Online. Association for Computational Linguistics.

\bibitem[{Kryscinski et~al.(2020{\natexlab{b}})Kryscinski, McCann, Xiong, and Socher}]{emnlp20-summ-fact}
Wojciech Kryscinski, Bryan McCann, Caiming Xiong, and Richard Socher. 2020{\natexlab{b}}.
\newblock \href {https://doi.org/10.18653/v1/2020.emnlp-main.750} {Evaluating the factual consistency of abstractive text summarization}.
\newblock In \emph{Proceedings of the 2020 Conference on Empirical Methods in Natural Language Processing, {EMNLP} 2020}, pages 9332--9346.

\bibitem[{Kwiatkowski et~al.(2019)Kwiatkowski, Palomaki, Redfield, Collins, Parikh, Alberti, Epstein, Polosukhin, Devlin, Lee, Toutanova, Jones, Kelcey, Chang, Dai, Uszkoreit, Le, and Petrov}]{kwiatkowski-etal-2019-natural}
Tom Kwiatkowski, Jennimaria Palomaki, Olivia Redfield, Michael Collins, Ankur Parikh, Chris Alberti, Danielle Epstein, Illia Polosukhin, Jacob Devlin, Kenton Lee, Kristina Toutanova, Llion Jones, Matthew Kelcey, Ming-Wei Chang, Andrew~M. Dai, Jakob Uszkoreit, Quoc Le, and Slav Petrov. 2019.
\newblock \href {https://doi.org/10.1162/tacl_a_00276} {Natural questions: A benchmark for question answering research}.
\newblock \emph{Transactions of the Association for Computational Linguistics}, 7:452--466.

\bibitem[{Lee et~al.(2023)Lee, Ping, Xu, Patwary, Fung, Shoeybi, and Catanzaro}]{lee2023factuality}
Nayeon Lee, Wei Ping, Peng Xu, Mostofa Patwary, Pascale Fung, Mohammad Shoeybi, and Bryan Catanzaro. 2023.
\newblock \href {http://arxiv.org/abs/2206.04624} {Factuality enhanced language models for open-ended text generation}.

\bibitem[{Lewis et~al.(2020)Lewis, Perez, Piktus, Petroni, Karpukhin, Goyal, K{\"{u}}ttler, Lewis, Yih, Rockt{\"{a}}schel, Riedel, and Kiela}]{nips20-rag}
Patrick S.~H. Lewis, Ethan Perez, Aleksandra Piktus, Fabio Petroni, Vladimir Karpukhin, Naman Goyal, Heinrich K{\"{u}}ttler, Mike Lewis, Wen{-}tau Yih, Tim Rockt{\"{a}}schel, Sebastian Riedel, and Douwe Kiela. 2020.
\newblock \href {https://proceedings.neurips.cc/paper/2020/hash/6b493230205f780e1bc26945df7481e5-Abstract.html} {Retrieval-augmented generation for knowledge-intensive {NLP} tasks}.
\newblock In \emph{Advances in Neural Information Processing Systems 33: Annual Conference on Neural Information Processing Systems 2020, NeurIPS 2020}.

\bibitem[{Li et~al.(2023)Li, Cheng, Zhao, Nie, and Wen}]{halueval}
Junyi Li, Xiaoxue Cheng, Wayne~Xin Zhao, Jian{-}Yun Nie, and Ji{-}Rong Wen. 2023.
\newblock \href {https://doi.org/10.48550/arXiv.2305.11747} {Halueval: {A} large-scale hallucination evaluation benchmark for large language models}.
\newblock \emph{CoRR}, abs/2305.11747.

\bibitem[{Li et~al.(2022)Li, Zhao, Lyu, and Wang}]{emnlp22-kgc}
Yanyang Li, Jianqiao Zhao, Michael~R. Lyu, and Liwei Wang. 2022.
\newblock \href {https://aclanthology.org/2022.emnlp-main.721} {Eliciting knowledge from large pre-trained models for unsupervised knowledge-grounded conversation}.
\newblock In \emph{Proceedings of the 2022 Conference on Empirical Methods in Natural Language Processing, {EMNLP} 2022}, pages 10551--10564.

\bibitem[{Liu et~al.(2023{\natexlab{a}})Liu, Zhang, and Liang}]{liu2023evaluating}
Nelson~F. Liu, Tianyi Zhang, and Percy Liang. 2023{\natexlab{a}}.
\newblock \href {http://arxiv.org/abs/2304.09848} {Evaluating verifiability in generative search engines}.

\bibitem[{Liu et~al.(2023{\natexlab{b}})Liu, Fabbri, Liu, Radev, and Cohan}]{gpt_sum}
Yixin Liu, Alexander~R. Fabbri, Pengfei Liu, Dragomir Radev, and Arman Cohan. 2023{\natexlab{b}}.
\newblock \href {https://doi.org/10.48550/arXiv.2305.14239} {On learning to summarize with large language models as references}.
\newblock \emph{CoRR}, abs/2305.14239.

\bibitem[{Liu et~al.(2022)Liu, Patwary, Prenger, Prabhumoye, Ping, Shoeybi, and Catanzaro}]{liu2022multistage}
Zihan Liu, Mostofa Patwary, Ryan Prenger, Shrimai Prabhumoye, Wei Ping, Mohammad Shoeybi, and Bryan Catanzaro. 2022.
\newblock \href {http://arxiv.org/abs/2203.08745} {Multi-stage prompting for knowledgeable dialogue generation}.

\bibitem[{Manakul et~al.(2023)Manakul, Liusie, and Gales}]{self_consistency}
Potsawee Manakul, Adian Liusie, and Mark J.~F. Gales. 2023.
\newblock \href {https://doi.org/10.48550/arXiv.2303.08896} {Selfcheckgpt: Zero-resource black-box hallucination detection for generative large language models}.
\newblock \emph{CoRR}, abs/2303.08896.

\bibitem[{Maynez et~al.(2020)Maynez, Narayan, Bohnet, and McDonald}]{acl20-summ-fact}
Joshua Maynez, Shashi Narayan, Bernd Bohnet, and Ryan~T. McDonald. 2020.
\newblock \href {https://doi.org/10.18653/v1/2020.acl-main.173} {On faithfulness and factuality in abstractive summarization}.
\newblock In \emph{Proceedings of the 58th Annual Meeting of the Association for Computational Linguistics, {ACL} 2020}, pages 1906--1919.

\bibitem[{Min et~al.(2023)Min, Krishna, Lyu, Lewis, tau Yih, Koh, Iyyer, Zettlemoyer, and Hajishirzi}]{min2023factscore}
Sewon Min, Kalpesh Krishna, Xinxi Lyu, Mike Lewis, Wen tau Yih, Pang~Wei Koh, Mohit Iyyer, Luke Zettlemoyer, and Hannaneh Hajishirzi. 2023.
\newblock \href {http://arxiv.org/abs/2305.14251} {Factscore: Fine-grained atomic evaluation of factual precision in long form text generation}.

\bibitem[{Nogueira et~al.(2019)Nogueira, Yang, Cho, and Lin}]{bert_ranker}
Rodrigo~Frassetto Nogueira, Wei Yang, Kyunghyun Cho, and Jimmy Lin. 2019.
\newblock \href {http://arxiv.org/abs/1910.14424} {Multi-stage document ranking with {BERT}}.
\newblock \emph{CoRR}, abs/1910.14424.

\bibitem[{OpenAI(2023)}]{gpt4}
OpenAI. 2023.
\newblock \href {https://doi.org/10.48550/arXiv.2303.08774} {{GPT-4} technical report}.
\newblock \emph{CoRR}, abs/2303.08774.

\bibitem[{Ouyang et~al.(2022)Ouyang, Wu, Jiang, Almeida, Wainwright, Mishkin, Zhang, Agarwal, Slama, Ray, Schulman, Hilton, Kelton, Miller, Simens, Askell, Welinder, Christiano, Leike, and Lowe}]{ouyang2022training}
Long Ouyang, Jeff Wu, Xu~Jiang, Diogo Almeida, Carroll~L. Wainwright, Pamela Mishkin, Chong Zhang, Sandhini Agarwal, Katarina Slama, Alex Ray, John Schulman, Jacob Hilton, Fraser Kelton, Luke Miller, Maddie Simens, Amanda Askell, Peter Welinder, Paul Christiano, Jan Leike, and Ryan Lowe. 2022.
\newblock \href {http://arxiv.org/abs/2203.02155} {Training language models to follow instructions with human feedback}.

\bibitem[{Pagnoni et~al.(2021)Pagnoni, Balachandran, and Tsvetkov}]{pagnoni-etal-2021-understanding}
Artidoro Pagnoni, Vidhisha Balachandran, and Yulia Tsvetkov. 2021.
\newblock \href {https://doi.org/10.18653/v1/2021.naacl-main.383} {Understanding factuality in abstractive summarization with {FRANK}: A benchmark for factuality metrics}.
\newblock In \emph{Proceedings of the 2021 Conference of the North American Chapter of the Association for Computational Linguistics: Human Language Technologies}, pages 4812--4829, Online. Association for Computational Linguistics.

\bibitem[{Pan et~al.(2023)Pan, Wu, Lu, Luu, Wang, Kan, and Nakov}]{pan2023factchecking}
Liangming Pan, Xiaobao Wu, Xinyuan Lu, Anh~Tuan Luu, William~Yang Wang, Min-Yen Kan, and Preslav Nakov. 2023.
\newblock \href {http://arxiv.org/abs/2305.12744} {Fact-checking complex claims with program-guided reasoning}.

\bibitem[{Peng et~al.(2023)Peng, Galley, He, Cheng, Xie, Hu, Huang, Liden, Yu, Chen, and Gao}]{peng2023check}
Baolin Peng, Michel Galley, Pengcheng He, Hao Cheng, Yujia Xie, Yu~Hu, Qiuyuan Huang, Lars Liden, Zhou Yu, Weizhu Chen, and Jianfeng Gao. 2023.
\newblock \href {http://arxiv.org/abs/2302.12813} {Check your facts and try again: Improving large language models with external knowledge and automated feedback}.

\bibitem[{Petroni et~al.(2021)Petroni, Piktus, Fan, Lewis, Yazdani, De~Cao, Thorne, Jernite, Karpukhin, Maillard et~al.}]{petroni2021kilt}
Fabio Petroni, Aleksandra Piktus, Angela Fan, Patrick Lewis, Majid Yazdani, Nicola De~Cao, James Thorne, Yacine Jernite, Vladimir Karpukhin, Jean Maillard, et~al. 2021.
\newblock Kilt: a benchmark for knowledge intensive language tasks.
\newblock In \emph{Proceedings of the 2021 Conference of the North American Chapter of the Association for Computational Linguistics: Human Language Technologies}, pages 2523--2544.

\bibitem[{Radford et~al.(2019)Radford, Wu, Child, Luan, Amodei, Sutskever et~al.}]{gpt2}
Alec Radford, Jeffrey Wu, Rewon Child, David Luan, Dario Amodei, Ilya Sutskever, et~al. 2019.
\newblock Language models are unsupervised multitask learners.
\newblock \emph{OpenAI blog}, 1(8):9.

\bibitem[{Raffel et~al.(2020)Raffel, Shazeer, Roberts, Lee, Narang, Matena, Zhou, Li, and Liu}]{raffel2020exploring}
Colin Raffel, Noam Shazeer, Adam Roberts, Katherine Lee, Sharan Narang, Michael Matena, Yanqi Zhou, Wei Li, and Peter~J. Liu. 2020.
\newblock \href {http://arxiv.org/abs/1910.10683} {Exploring the limits of transfer learning with a unified text-to-text transformer}.

\bibitem[{Rajpurkar et~al.(2016)Rajpurkar, Zhang, Lopyrev, and Liang}]{rajpurkar2016squad}
Pranav Rajpurkar, Jian Zhang, Konstantin Lopyrev, and Percy Liang. 2016.
\newblock \href {http://arxiv.org/abs/1606.05250} {Squad: 100,000+ questions for machine comprehension of text}.

\bibitem[{Santhanam et~al.(2021)Santhanam, Khattab, Saad{-}Falcon, Potts, and Zaharia}]{colbertv2}
Keshav Santhanam, Omar Khattab, Jon Saad{-}Falcon, Christopher Potts, and Matei Zaharia. 2021.
\newblock \href {http://arxiv.org/abs/2112.01488} {Colbertv2: Effective and efficient retrieval via lightweight late interaction}.
\newblock \emph{CoRR}, abs/2112.01488.

\bibitem[{Schuster et~al.(2021)Schuster, Fisch, and Barzilay}]{schuster-etal-2021-get}
Tal Schuster, Adam Fisch, and Regina Barzilay. 2021.
\newblock \href {https://doi.org/10.18653/v1/2021.naacl-main.52} {Get your vitamin {C}! robust fact verification with contrastive evidence}.
\newblock In \emph{Proceedings of the 2021 Conference of the North American Chapter of the Association for Computational Linguistics: Human Language Technologies}, pages 624--643, Online. Association for Computational Linguistics.

\bibitem[{Shuster et~al.(2021)Shuster, Poff, Chen, Kiela, and Weston}]{shuster-etal-2021-retrieval-augmentation}
Kurt Shuster, Spencer Poff, Moya Chen, Douwe Kiela, and Jason Weston. 2021.
\newblock \href {https://doi.org/10.18653/v1/2021.findings-emnlp.320} {Retrieval augmentation reduces hallucination in conversation}.
\newblock In \emph{Findings of the Association for Computational Linguistics: EMNLP 2021}, pages 3784--3803, Punta Cana, Dominican Republic. Association for Computational Linguistics.

\bibitem[{Somers(1962)}]{somers1962new}
Robert~H Somers. 1962.
\newblock A new asymmetric measure of association for ordinal variables.
\newblock \emph{American sociological review}, pages 799--811.

\bibitem[{Thorne et~al.(2018)Thorne, Vlachos, Christodoulopoulos, and Mittal}]{fever}
James Thorne, Andreas Vlachos, Christos Christodoulopoulos, and Arpit Mittal. 2018.
\newblock \href {https://doi.org/10.18653/v1/N18-1074} {{FEVER}: a large-scale dataset for fact extraction and {VER}ification}.
\newblock In \emph{Proceedings of the 2018 Conference of the North {A}merican Chapter of the Association for Computational Linguistics: Human Language Technologies, Volume 1 (Long Papers)}, pages 809--819, New Orleans, Louisiana. Association for Computational Linguistics.

\bibitem[{Touvron et~al.(2023)Touvron, Lavril, Izacard, Martinet, Lachaux, Lacroix, Rozière, Goyal, Hambro, Azhar, Rodriguez, Joulin, Grave, and Lample}]{touvron2023llama}
Hugo Touvron, Thibaut Lavril, Gautier Izacard, Xavier Martinet, Marie-Anne Lachaux, Timothée Lacroix, Baptiste Rozière, Naman Goyal, Eric Hambro, Faisal Azhar, Aurelien Rodriguez, Armand Joulin, Edouard Grave, and Guillaume Lample. 2023.
\newblock \href {http://arxiv.org/abs/2302.13971} {Llama: Open and efficient foundation language models}.

\bibitem[{Wadden et~al.(2020)Wadden, Lin, Lo, Wang, van Zuylen, Cohan, and Hajishirzi}]{wadden-etal-2020-fact}
David Wadden, Shanchuan Lin, Kyle Lo, Lucy~Lu Wang, Madeleine van Zuylen, Arman Cohan, and Hannaneh Hajishirzi. 2020.
\newblock \href {https://doi.org/10.18653/v1/2020.emnlp-main.609} {Fact or fiction: Verifying scientific claims}.
\newblock In \emph{Proceedings of the 2020 Conference on Empirical Methods in Natural Language Processing (EMNLP)}, pages 7534--7550, Online. Association for Computational Linguistics.

\bibitem[{Wang et~al.(2020)Wang, Cho, and Lewis}]{QGQA}
Alex Wang, Kyunghyun Cho, and Mike Lewis. 2020.
\newblock \href {https://doi.org/10.18653/v1/2020.acl-main.450} {Asking and answering questions to evaluate the factual consistency of summaries}.
\newblock In \emph{Proceedings of the 58th Annual Meeting of the Association for Computational Linguistics}, pages 5008--5020, Online. Association for Computational Linguistics.

\bibitem[{Wang et~al.(2023)Wang, Sun, Li, Ouyang, Wu, Zhang, Li, and Wang}]{gpt_ner}
Shuhe Wang, Xiaofei Sun, Xiaoya Li, Rongbin Ouyang, Fei Wu, Tianwei Zhang, Jiwei Li, and Guoyin Wang. 2023.
\newblock \href {https://doi.org/10.48550/arXiv.2304.10428} {{GPT-NER:} named entity recognition via large language models}.
\newblock \emph{CoRR}, abs/2304.10428.

\bibitem[{Wei et~al.(2022)Wei, Bosma, Zhao, Guu, Yu, Lester, Du, Dai, and Le}]{wei2022finetuned}
Jason Wei, Maarten Bosma, Vincent~Y. Zhao, Kelvin Guu, Adams~Wei Yu, Brian Lester, Nan Du, Andrew~M. Dai, and Quoc~V. Le. 2022.
\newblock \href {http://arxiv.org/abs/2109.01652} {Finetuned language models are zero-shot learners}.

\bibitem[{Xie et~al.(2023)Xie, Dong, Wang, Lv, Yao, Gan, Wu, Li, Li, Liu, and Ma}]{xie2023t2ranking}
Xiaohui Xie, Qian Dong, Bingning Wang, Feiyang Lv, Ting Yao, Weinan Gan, Zhijing Wu, Xiangsheng Li, Haitao Li, Yiqun Liu, and Jin Ma. 2023.
\newblock \href {http://arxiv.org/abs/2304.03679} {T2ranking: A large-scale chinese benchmark for passage ranking}.

\bibitem[{Yu et~al.(2023)Yu, Iter, Wang, Xu, Ju, Sanyal, Zhu, Zeng, and Jiang}]{yu2023generate}
Wenhao Yu, Dan Iter, Shuohang Wang, Yichong Xu, Mingxuan Ju, Soumya Sanyal, Chenguang Zhu, Michael Zeng, and Meng Jiang. 2023.
\newblock \href {http://arxiv.org/abs/2209.10063} {Generate rather than retrieve: Large language models are strong context generators}.

\bibitem[{Zhang et~al.(2023)Zhang, Li, Cui, Cai, Liu, Fu, Huang, Zhao, Zhang, Chen, Wang, Luu, Bi, Shi, and Shi}]{zhang2023sirens}
Yue Zhang, Yafu Li, Leyang Cui, Deng Cai, Lemao Liu, Tingchen Fu, Xinting Huang, Enbo Zhao, Yu~Zhang, Yulong Chen, Longyue Wang, Anh~Tuan Luu, Wei Bi, Freda Shi, and Shuming Shi. 2023.
\newblock \href {http://arxiv.org/abs/2309.01219} {Siren's song in the ai ocean: A survey on hallucination in large language models}.

\end{thebibliography}
\bibliographystyle{acl_natbib}

\appendix

\section*{Appendix}

\begin{figure}[!t]
    \centering
     \includegraphics[width=0.48\textwidth]{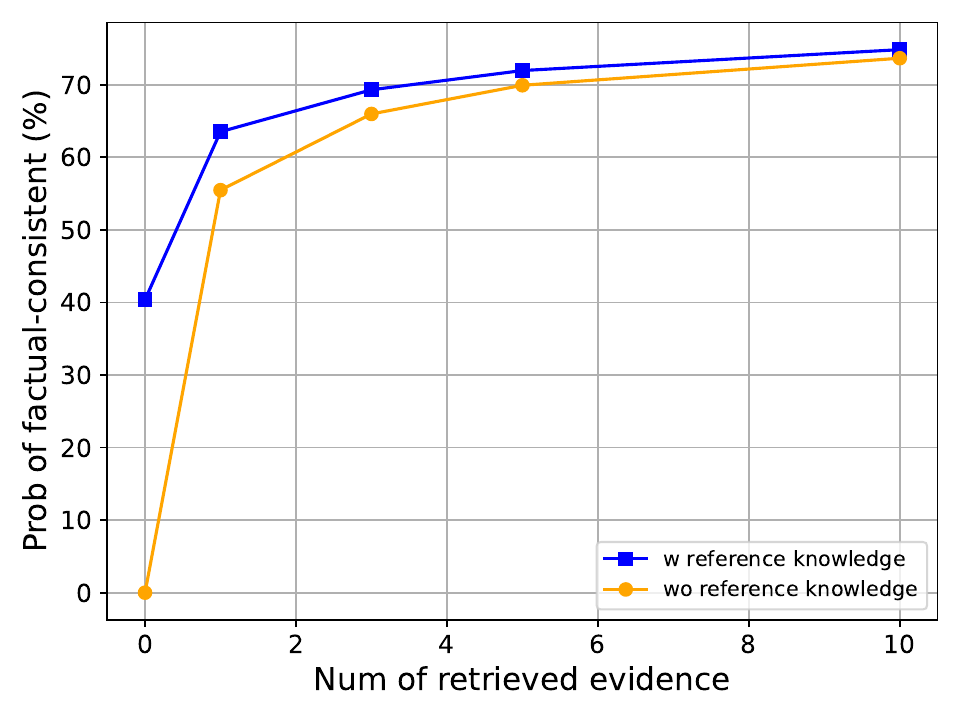}
    \caption{The influence of reference knowledge in factuality evaluation weakens as the amount of retrieved evidence increases.}
    \label{fig:knowledge_impact}
\end{figure}

\section{Details of Problem Formulation}
\label{app:problem}
We provide a formulation of the two-step process for knowledge-intensive tasks, as illustrated in Fig.\ref{fig:framework}.
Formally, the knowledge-intensive generation problem can be formulated as the following chain rule:
\begin{equation}\small
P(a | q, \mathcal{K})= \sum\nolimits_{k}P( k | q, \mathcal{K} ) P( a | q, k )
\label{dialog_persona}
\end{equation}
where $P( k | q, \mathcal{K} )$ is the knowledge acquisition process and $P( a | q, k )= \prod\nolimits_{t=1}^{N} P( a_t | a_{1:t-1}, q, k)$ is the autoregressive answer generation process based on the acquired knowledge. 

Retrieval-based knowledge acquisition methods use a retrieval model to retrieve the most relevant knowledge from the knowledge resource $\mathcal{K} = \{d_1, d_2, \ldots, d_{K}\}$ composed of $K$ documents:
\begin{equation}\small
P(k=d_i | q, \mathcal{K})= \frac{ e^{\mathrm{sim}(q, d_i)} }{\sum_{j=1}^{K}{e^{\mathrm{sim}(q, d_j)}}}
\end{equation}
where $\mathrm{sim}(\cdot)$ function is used to measure the similarity, e.g., cosine similarity, between the query and the knowledge document. 

Generation-based knowledge acquisition methods prompt a large language model to directly generate the required knowledge:
\begin{equation}\small
P(k | q, \mathcal{K})= \prod\nolimits_{t=1}^{M} P_{\mathcal{K}}( k_t | k_{1:t-1}, q, \texttt{prompt} )
\end{equation}
where \texttt{prompt} denotes the zero-shot or few-shot prompt and the LLM is regarded as the knowledge resource $\mathcal{K}$ and $P_{\mathcal{K}}$ stands for the distribution induced by the LLM.

\begin{table*}[t!]
\centering
\small
\begin{adjustbox}{max width=\textwidth}
\begin{tabular}{l|c|c}
\toprule
Dataset & Prompts & Best \\
\midrule
\multirow{4}{*}{\textsc{NQ}} & Topic: \{topic\} $\backslash$n Query: \{query\} $\backslash$n Related wikipedia knowledge: &  \\
 & Topic: \{topic\} $\backslash$n Generate a background document from Wikipedia to answer the given question. $\backslash$n \{query\} $\backslash$n &  \\
 & Topic: \{topic\} $\backslash$n Generate a Wikipedia knowledge to answer the given question.$\backslash$n Question: \{query\} $\backslash$n Wikipedia knowledge: &  \\
 & Topic: \{topic\} $\backslash$n Generate a Wikipedia to answer the given question.$\backslash$n Question: \{query\} $\backslash$n Wikipedia: & \checkmark \\

\midrule
\multirow{4}{*}{\textsc{WoW}} & Topic: \{topic\} $\backslash$n Query: \{utterance\} $\backslash$n Related Wikipedia knowledge: &  \\
 & Topic: \{topic\} $\backslash$n Generate a background document from Wikipedia to reply to the utterance. $\backslash$n \{utterance\} $\backslash$n &  \\
 & Topic: \{topic\} $\backslash$n Generate a Wikipedia knowledge to answer the given question.$\backslash$n Utterance: \{utterance\} $\backslash$n Wikipedia knowledge: &  \\
 & Topic: \{topic\} $\backslash$n Generate a Wikipedia to answer the given question.$\backslash$n Question: \{utterance\} $\backslash$n Wikipedia: & \checkmark \\

\bottomrule
\end{tabular}
\end{adjustbox}
\caption{List of human prompts we tried for zero-shot knowledge generation, evaluated on the validation set of NQ, WoW. \{\} represents placeholder, and 'utterance' denotes the last utterance of the dialogue partner. We use \checkmark to denote the prompt achieving the best performance.}
\vspace{-0.1in}
\vspace{0.2in}
\label{tab:human-prompt-zero-shot}
\end{table*}
\begin{table*}[t!]
\centering
\small
\begin{adjustbox}{max width=\textwidth}
\begin{tabular}{l|c|c}
\toprule
Dataset & Prompts & Best \\
\midrule
\multirow{4}{*}{\textsc{NQ}} & Topic: \{topic\} $\backslash$n Query: \{query\} $\backslash$n Related Wikipedia knowledge: \{knowledge\} & \checkmark \\
 & Topic: \{topic\} $\backslash$n Query: \{query\} $\backslash$n Knowledge: \{knowledge\} & \\
 & Topic: \{topic\} $\backslash$n Query: \{query\} $\backslash$n Document: \{knowledge\} & \\
 & Topic: \{topic\} $\backslash$n Generate a background document from Wikipedia to answer the given question. $\backslash$n \{query\} $\backslash$n \{knowledge\} &  \\
 & Topic: \{topic\} $\backslash$n Generate a Wikipedia to answer the given question.$\backslash$n Question: \{query\} $\backslash$n Wikipedia: \{knowledge\} & \\

\midrule
\multirow{4}{*}{\textsc{WoW}} & Topic: \{topic\} $\backslash$n Query: \{utterance\} $\backslash$n Related Wikipedia knowledge: \{knowledge\} & \checkmark  \\
 & Topic: \{topic\} $\backslash$n Query: \{utterance\} $\backslash$n Knowledge: \{knowledge\} & \\
 & Topic: \{topic\} $\backslash$n Query: \{utterance\} $\backslash$n Document: \{knowledge\} & \\
 & Topic: \{topic\} $\backslash$n Generate a background document from Wikipedia to reply to the utterance. $\backslash$n \{utterance\} $\backslash$n \{knowledge\} &  \\
 & Topic: \{topic\} $\backslash$n Generate a Wikipedia to answer the given question.$\backslash$n Question: \{utterance\} $\backslash$n Wikipedia: \{knowledge\} & \\

\bottomrule
\end{tabular}
\end{adjustbox}
\caption{List of example templates we tried for few-shot knowledge generation.}
\vspace{-0.1in}
\vspace{0.2in}
\label{tab:fewshot-knowledge-prompt}
\end{table*}
\begin{table*}[t!]
\centering
\small
\begin{adjustbox}{max width=\textwidth}
\begin{tabular}{l|c|c}
\toprule
Dataset & Prompts & Best \\
\midrule
\multirow{4}{*}{\textsc{NQ}} & Topic: \{topic\} $\backslash$n Passage: \{knowledge\} $\backslash$n Query: \{query\} $\backslash$n Answer: \{answer\} & \checkmark \\
 & Topic: \{topic\} $\backslash$n Read the passage and answer the question below:$\backslash$n Passage: \{knowledge\} $\backslash$n Question: \{query\} $\backslash$n Answer: \{answer\} & \\
 & Topic: \{topic\} $\backslash$n Using the knowledge from the passage to answer the question below:$\backslash$n Passage: \{knowledge\} $\backslash$n Question: \{query\} $\backslash$n Answer: \{answer\} & \\

\midrule
\multirow{4}{*}{\textsc{WoW}} & Topic: \{topic\} $\backslash$n Passage: \{knowledge\} $\backslash$n Speaker 1: \{utterance\} $\backslash$n Speaker 2: \{response\} & \checkmark  \\
 &  Topic: \{topic\} $\backslash$n Knowledge: \{knowledge\} $\backslash$n Speaker 1: \{utterance\} $\backslash$n Speaker 2: \{response\} & \\
 &  Topic: \{topic\} $\backslash$n Grounding document: \{knowledge\} $\backslash$n Speaker 1: \{utterance\} $\backslash$n Speaker 2: \{response\} & \\
 &  Passage: \{knowledge\} $\backslash$n Query: \{utterance\} $\backslash$n Answer: \{response\} & \\
  & Topic: \{topic\} $\backslash$n Using the knowledge from the passage, complete the dialogue below: \{knowledge\} $\backslash$n Speaker 1: \{utterance\} $\backslash$n Speaker 2: \{response\} & \\

\bottomrule
\end{tabular}
\end{adjustbox}
\caption{List of example templates we tried for few-shot answer generation. }
\vspace{-0.1in}
\vspace{0.2in}
\label{tab:fewshot-answer-prompt}
\end{table*}

\section{Analysis of Reference Knowledge}\label{app:ground-truth}

We investigated the importance of reference knowledge in evaluating the factuality of generated knowledge. Specifically, we conducted FLAN-T5 experiments on the WoW dataset using a zero-shot approach. Two sets of experiments were performed: one included reference knowledge in the retrieved evidence pool, while the other did not. Figure~\ref{fig:knowledge_impact} illustrates our findings, indicating that the group with reference knowledge exhibits a clear advantage when the number of retrieved evidence is limited. However, as the number of retrieved evidence increases, the performance of both groups converges. These results suggest that reference knowledge is dispensable, particularly when a significant amount of evidence is available. When the number of retrieved evidence surpasses ten, the impact of reference knowledge becomes negligible. We hope this will provide valuable insights for future designs of factuality assessment for generated knowledge.

\section{Details of Baselines}\label{app:baseline}

\noindent
\textbf{DPR}~\cite{karpukhin2020dense} is a supervised dense retrieval model trained on several QA datasets (including NQ) to retrieve the most relevant Wikipedia passages given a query.

\noindent
\textbf{FLAN-T5}~\cite{wei2022finetuned} is an enhanced version of T5~\cite{raffel2020exploring} that is instruction-finetuned in 1.8k NLP datasets to acquire the generalization ability to unseen tasks.

\noindent
\textbf{LLaMA}~\cite{touvron2023llama} is an open-source foundation language model trained on publicly available datasets and shows competitive performance with the best models, including GPT-3 (175B) and PaLM-540B.

\noindent
\textbf{ChatGPT} is a sibling model to InstructGPT \cite{ouyang2022training} that is trained to follow instructions in a prompt and provide a detailed response. 
We adopt \texttt{text-davinci-003} version for evaluation.

\section{Details of Datasets}\label{app:dataset}
\noindent
\textbf{Natural Questions (NQ)}~\cite{kwiatkowski-etal-2019-natural} is an open-domain QA dataset, where the questions are mined from real Google search queries. The corresponding ground truth knowledge and the answers to the questions are paragraphs and short spans in the Wikipedia pages.

\noindent
\textbf{Wizard of Wikipedia (WoW)}~\cite{dinan2018wizard} is a knowledge-grounded dialogue dataset designed for information-seeking scenarios, where one speaker introduces knowledge related to a topic to the other speaker by grounding his/her responses in a specific sentence from a Wikipedia page.

\section{Implementation Details} \label{app:implement}
\begin{table*}[t!]
\centering
\small
\begin{adjustbox}{max width=\textwidth}
\begin{tabular}{l|c|c}
\toprule
Metric & Model & Link \\
\midrule
\multirow{2}{*}{\texttt{Factuality}} & NLI-RoBERTa-large & \url{https://huggingface.co/sentence-transformers/nli-roberta-large} \\
 & ColBERTv2 & \url{https://github.com/stanford-futuredata/ColBERT} \\

\midrule
\multirow{1}{*}{\texttt{Relevance}} & BERT-{ranking}-large & \url{https://github.com/nyu-dl/dl4marco-bert} \\

\midrule
\multirow{2}{*}{\texttt{Coherence}} & GPT-neo-2.7B & \url{https://huggingface.co/EleutherAI/gpt-neo-2.7B} \\
 & Coherence-Momentum  & \url{https://huggingface.co/aisingapore/coherence-momentum} \\

\midrule
\multirow{1}{*}{\texttt{Informativeness}} & GPT-neo-2.7B & \url{https://huggingface.co/EleutherAI/gpt-neo-2.7B} \\

\midrule
\multirow{1}{*}{\texttt{Helpfulness}} & LLaMA-65B & \url{https://github.com/facebookresearch/llama/tree/main} \\

\midrule
\multirow{2}{*}{\texttt{Validity}} & NLI-RoBERTa-large & \url{https://huggingface.co/sentence-transformers/nli-roberta-large} \\
 & ColBERTv2 & \url{https://github.com/stanford-futuredata/ColBERT} \\

\bottomrule
\end{tabular}
\end{adjustbox}
\caption{List of all models that we use in designing our framework. }
\vspace{-0.1in}
\vspace{0.2in}
\label{tab:metric_model}
\end{table*}

All the metrics we designed are model-based metrics, utilizing solely off-the-shelf models. We present the models we used in Table \ref{tab:metric_model}.

\section{Prompts for Knowledge Generation}
\label{sec:knowledge-gen-prompt}

\subsection{Zero-shot Prompts}
In our experiments, we observed that zero-shot prompting was highly unstable. Therefore, we conduct experiments using multiple human prompts and select the most effective ones for the WoW and NQ datasets. The human prompts we evaluate are listed in Table~\ref{tab:human-prompt-zero-shot}.

\subsection{Few-shot Prompts}

In the few-shot setting, our prompt is constructed using k randomly chosen examples from the training set:
\begin{equation*} \small
    prompt=(\textit{example}_1 \verb!\n! ~ ... ~ \textit{example}_k \verb!\n! ~ \textit{example}_{test})
\end{equation*}
The example templates utilized for knowledge generation are provided in Table \ref{tab:fewshot-knowledge-prompt}. 
Please note that $\textit{example}_{test}$ differs from $\textit{example}_{i}$ as it does not contain $\{knowledge\}$ in the placeholder.

\section{Prompts for Answer Generation}
\label{sec:answer-gen-prompt}

We adopt few-shot prompting on the LLaMA model in answer generation and the example templates used for answer generation are provided in Table \ref{tab:fewshot-answer-prompt}.

\section{Detailed Correlations between Intrinsic and Extrinsic Metrics}
\label{sec:break_down}




 



\begin{table}[t!]
\setlength{\abovecaptionskip}{5pt}   
\setlength{\belowcaptionskip}{0pt}
\setlength{\tabcolsep}{0.2mm}
\small
\centering
\begin{tabular}{llccccc}
\toprule
\multirow{2}{*}{\bf Model} & \multirow{2}{*}{\bf Extrinsic} & \multicolumn{5}{c}{\bf Instrinsic} \\
\cmidrule(lr){3-7}
& & Fact. & Rel. & Coh-{sent.} & Coh-{para.} & Info. \\

\midrule
\multirow{2}{*}{\textsc{FLAN-T5}} & helpful. & 0.15$^\dagger$ & -0.21$^\dagger$ & 0.20$^\dagger$ & -0.21$^\dagger$ & 0.02 \\
 & validity & 0.23$^\dagger$ & -0.16$^\dagger$ & 0.14$^\dagger$ & -0.10$^\dagger$ & 0.07 \\

\midrule
\multirow{2}{*}{\textsc{LLaMA}} & helpful.  & 0.03 & 0.05 & 0.06 & -0.09$^\dagger$ & -0.01 \\
 & validity  & 0.09$^\dagger$ & 0.07 & 0.05 & -0.06 & -0.03 \\

\midrule
\multirow{2}{*}{\textsc{ChatGPT}} & helpful. & 0.16$^\dagger$ & 0.03 & 0.08 & 0.02 & -0.04$^\dagger$ \\
 & validity & 0.22$^\dagger$ & 0.13$^\dagger$ & 0.02$^\dagger$ & 0.09$^\dagger$ & 0.03 \\

\bottomrule
\vspace{-2mm}
\end{tabular}

\caption{\label{tab:inter_correlation_nq_breakdowns} The Somers' correlation between intrinsic and extrinsic metrics in zero-shot setting on NQ. Correlation scores with $p\text{-}value < 0.05$ are marked with $^\dagger$. 
}
\vspace{-1mm}
\end{table}




 


We listed the detailed correlations between intrinsic and extrinsic metrics for LLaMA, FLAN-T5, and ChatGPT on the NQ dataset in Table \ref{tab:inter_correlation_nq_breakdowns}.

\section{Table of Model Size Impact}
\label{sec:model_size}
\begin{table}[t!]
\setlength{\tabcolsep}{0.54mm}
\small
\centering
\begin{tabular}{l|l|c|c|c|c|c|c}
\toprule

Model & Size & Fact. & Rel. & Coh. & Info & Help. & Val. \\

\midrule
\multirow{3}{*}{\bf LLaMA} & 65B & \textbf{0.942} & \textbf{0.732} & \textbf{0.824} & \textbf{0.757} & \textbf{0.219} & \textbf{0.420} \\

 & 33B & 0.656 & \underline{0.633} & 0.734 & 0.608 & \underline{0.203} & \underline{0.402} \\
 
 & 7B & \underline{0.773} & 0.626 & \underline{0.805} & \underline{0.662} & 0.154 & 0.375 \\

\midrule
\multirow{3}{*}{\bf FLAN-T5} & 11B & \underline{0.584} & \underline{0.685} & 0.778 & 0.673 & \textbf{-0.146} & \textbf{0.325} \\

 & 3B & \textbf{0.657} & 0.663 & \textbf{0.816} & \underline{0.708} & \underline{-0.155} & \underline{0.324} \\
 
 & 780M & 0.506 & \textbf{0.729} & \underline{0.793} & \textbf{0.729} & -0.162 & 0.252 \\

\bottomrule
\end{tabular}

\caption{\label{tab:model_scale_num} Performance on NQ with varying sizes of FLAN-T5 and LLaMA as knowledge generators.  
The max(0, .) operation in Eq.\ref{eq:helpful} has been excluded to emphasize the sequential relationship among different sizes of FLAN-T5.
 \textbf{Bold} and \underline{Underlined} results represent the best and second-best performances for each model, respectively.
}

\end{table}

We list the specific numerical values of performance scaling with the model size in Table \ref{tab:model_scale_num}, including LLaMA-65B/33B/7B and FLAN-T5-11B/3B/780M.

\begin{table*}[t!]
\centering
{\small
    \begin{adjustbox}{max width=\textwidth}
\begin{tabular}{l|c|l}
\toprule
Dimension & Value & Description \\
\midrule
 & 2 & All sentences in $k$ are factually accurate and the information in them can be verified with reliable evidence. \\
Factuality& 1 & $k$ contains at least one sentence with non-verified information, while others are factually accurate. \\
& 0 & $k$ contains at least one sentence with at least one factual error that is inconsistent with reliable knowledge. \\
\midrule
 & 2 & $k$ is highly relevant to the topic and query/utterance. \\
Relevance& 1 & $k$ is relevant to the topic but less relevant to the query/utterance. \\
& 0 & $k$ is irrelevant to both the topic and query/utterance. \\
\midrule
 & 2 & $k$ is very coherent and fluent (do not consider the truncation at the end due to the maximum generation length). \\
Coherence& 1 & $k$ has some minor incoherence or lack of fluency, \textit{e.g.}, phrase or sentence repetition, but it does not affect understanding. \\
& 0 & $k$ has significant coherence and fluency issues that are hard to understand. \\
\midrule
 & 2 & $k$ contains informative content that you don't know before. \\
Informative & 1 & $k$ contains limited or trivial information against your knowledge. \\
& 0 & $k$ fails to provide any meaningful information. \\
\midrule
 & 2 & $k$ directly provides or contains the correct answer. \\
Helpfulness& 1 & $k$ indirectly help in generating the correct answer. \\
& 0 & $k$ does not contain any useful information for the correct answer.\\
\midrule
 & 2 & The answer generated based on $k$ is correct. \\
Validity& 1 & The correctness of the generated answer cannot be determined. \\
& 0 & The answer generated based on $k$ is completely incorrect. \\
\bottomrule
\end{tabular}
\end{adjustbox}

    }
    \caption{Annotation guideline of LLM generated knowledge. }
    \label{tab:human_instruct}
\end{table*}

\begin{table}[t!]
\setlength{\tabcolsep}{2.0mm}
\small
\centering
\begin{tabular}{l|cccccc}
\toprule

Model & Fact. & Rel. & Coh. & Info. & Help. & Val.\\
\midrule
ChatGPT & 0.71 & 0.57 & 0.52 & 0.40 & 0.79 & 0.54 \\
\bottomrule
\end{tabular}

\caption{\label{tab:human_col_wow} Somer's D correlation of metrics with the human annotation on WoW ).
$p$-$value$ for all results are < 0.05.
We report the maximum for coherence. 
}

\vspace{-1mm}
\end{table}

\section{Details of Human Evaluation}
\label{sec:detail_human_eval}

\subsection{Human Annotation}
\label{sec:label_instruct}

We conducted a human evaluation with 400 samples from the NQ and WoW test set. 
Among these, 320 samples were from the zero-shot setting in the NQ dataset, involving all four models, while 80 samples were from the few-shot setting in the WoW dataset, involving one model (ChatGPT). 
Three expert annotators, who were familiar with the tasks, were employed to rate the acquired knowledge and generated answers based on four intrinsic perspectives and two extrinsic perspectives. Each perspective was scored on a scale of 0, 1, or 2, representing unacceptable, acceptable, and excellent, respectively. The average kappa value of the annotation is 0.612 on 20\% cross-annotation data.
The detailed instructions for the human annotation can be found in Table \ref{tab:human_instruct}. 

Note for factuality assessment, the reliable evidence for the generated knowledge $k$ is acquired by the following process: For each sentence in $k$, we use it as the query to search Google, and 
regard the top1 Wikipedia webpage as a reliable knowledge source to verify the factuality of this sentence. Another point worth noting is that for the evaluation of validity in WoW, we reused the factuality evaluation process since the responses in WoW are open-ended.

\subsection{Human Evaluation Results on WoW}
\label{app:wow_human}

Based on the provided annotations, we assessed the correlation between ChatGPT's automatic metrics and human judgement on the WoW dataset. The results are presented in Table \ref{tab:human_col_wow}.

\subsection{Baseline Metrics} \label{sec:bl_metrics}
We compared it with three reference-reliant metrics in knowledge evaluation. Their definitions and calculation methods are as follows:

\noindent
\textbf{Hallucinated NE Ratio (HE)} 
\cite{lee2023factuality} proposed a NE-based metric that is designed with an intuition that a model is hallucinating (making factual errors) if it generates an NE that does not appear in the reference knowledge source. The NE-based metric can be calculated as:
$
\textsc{HNE} = {|\textsc{Hallu}_{\text{NE}}|} ~/~ {|\textsc{All}_{\text{NE}}|}
$
where $\textsc{All}_{\text{NE}}$ is the set of all the NEs detected in the LM generation, and $\textsc{Hallu}_{\text{NE}}$ is a subset of $\textsc{NE}_{\text{All}}$ that does not appear in the reference Wikipedia knowledge. Note that evaluating \halluNE~requires the existence of reference knowledge. We adopt $-\textsc{HNE}$ when computing the correlation with human judgement.

\noindent
\textbf{Entailment Ratio (ER)} \cite{lee2023factuality} also introduces an NLI-based approach to assess factual knowledge by measuring its entailing relationship with ground-truth/reference knowledge. The entailment ratio is computed as follows: 
\entailedRatio $= | \textsc{Entail}_{\text{gen}}| ~/~ {|\textsc{All}_{\text{gen}}|},$
where $\textsc{All}_{\text{gen}}$ is a set of all generated knowledge, and $\textsc{Entail}_{\text{gen}}$ is the set of generated knowledge that can be entailed by the NLI model. Specifically, we use the entailment probability of each example as its ER score.

\noindent
\textbf{F1 of knowledge (F1)} \cite{liu2022multistage} employs a unigram F1 score to evaluate the quality of generated knowledge. This metric measures the overlap between the generated knowledge and the reference knowledge by evaluating word-level matches. By assessing the degree of agreement, the F1 metric provides an estimation of the knowledge quality, specifically from a relevance perspective.

\noindent
\textbf{NLI-weak-supervised} \cite{emnlp20-summ-fact} train a classification model on constructed data to perform consistency checking on (document, sentence) pairs. We chose the factCC version as our baseline.
 
\noindent
\textbf{NLI-decompose-claim} \cite{glover2022revisiting} found that in general, sentence-level decomposition is preferable for the hypothesis side of the NLI input. So we also decompose the generated knowledge into sentences and then aggregate the sentence-level scores to produce a document-level score.

\noindent
\textbf{NLI-multitask} fine-tunes the DeBERTa-v3-large model on FEVER and two NLI datasets.

\noindent
\textbf{Exact Match (EM)} \cite{rajpurkar2016squad} use Exact Match to measure the percentage of predictions that match its ground truth answers exactly.

\end{document}